\documentclass[10pt,conference]{IEEEtran}
\IEEEoverridecommandlockouts
\usepackage[utf8]{inputenc}
\usepackage[T1]{fontenc}
\usepackage{microtype}
\usepackage{amsmath,amssymb}
\usepackage{amsthm}
\usepackage{graphicx}
\usepackage{hyperref}
\usepackage{booktabs}
\usepackage{multirow}
\usepackage{url}
\usepackage{algorithm}
\usepackage[table]{xcolor}
\usepackage{algpseudocode}
\usepackage{threeparttable}
\usepackage{adjustbox}
\usepackage{cite}
\usepackage{setspace}
\usepackage{subcaption}
\usepackage{listings}
\usepackage{helvet}
\usepackage{tikz}
\usetikzlibrary{positioning, arrows.meta, shapes.geometric, calc, fit, backgrounds}

\newtheorem{definition}{Definition}

\title{DeepEdu-F: A Framework for Efficient and Scalable Agentic LLMs in Vietnamese Education}
\author{
    \IEEEauthorblockN{Quang Nguyen, Hieu Nguyen, Toan Pham, Hien Hoang, Cong Tran, Nam Vu}
    \IEEEauthorblockA{
        \textit{Posts and Telecommunications Institute of Technology, Ha Noi, Vietnam} \\
        Email: quangnm@ptit.edu.vn, \{HieuNPT.B23CE031, HienHC.B23CE028, \\ ToanPQ.B23CN835\}@stu.ptit.edu.vn, congtt@ptit.edu.vn
    }
}

\begin{document}
\raggedbottom

\maketitle

\bstctlcite{IEEEexample:BSTcontrol}

\begin{abstract}
Cloud AI assistants send student data to foreign servers, conflicting with data-sovereignty laws such as Vietnam's Decree 53, and are not grounded in the national curriculum, so they often hallucinate on local content. Self-hosting an open model keeps data on-premise but faces two costs that weight quantization does not remove. A tutoring session revisits the same long material every turn, so once prefix caching removes the prefill cost, decoding over a growing KV Cache dominates response time; and grounding the tutor in local knowledge by fine-tuning is too costly to repeat as that knowledge changes. We present \textbf{\mbox{DeepEdu-F}}, a framework for the Vietnamese AI tutor DeepEdu with two components. \textbf{LastSet} selects the critical KV Cache entries once per turn from the question's attention, decodes the whole turn from that set, and keeps the full cache so later questions can select again. \textbf{REPAIR} (Reliable Evidence-based Playbook Adaptation and Insight Revision) curates a verified playbook from past interactions instead of updating weights. LastSet decodes up to $3.1\times$ faster than dense decoding at a 256k-token context, faster than four sparse baselines, with accuracy close to full attention. Compared with ACE (ADD), REPAIR raises final accuracy from $70.0\%$ to $80.5\%$ on Formula and from $53.6\%$ to $59.7\%$ on FiNER, while mean task/scenario goal completion across AppWorld's normal and challenge splits increases from $9.28\%$ to $16.28\%$.

\end{abstract}

\begin{IEEEkeywords}
Large Language Models, AI Education, Long-Context Inference, Model Quantization, Data Sovereignty, Self-improving Agents
\end{IEEEkeywords}

\section{Introduction}

Large Language Models (LLMs) have moved beyond text generation into multimodal, tool-augmented agents capable of complex reasoning, planning, and personalized interaction \cite{touvron2023llama, bai2023qwen, yao2023react}. In education, this puts within reach something long unavailable in resource-constrained systems: a one-on-one tutor that adapts to each student's pace and cognitive profile \cite{llmeducation2026}. Our goal is \textbf{DeepEdu}: an AI-tutoring platform that gives Vietnamese students a more effective, curriculum-aligned learning path, with LLM agents as its foundation. Realizing this vision, however, runs into two obstacles that a generic chatbot does not solve.

A natural first question is why not simply use a cloud assistant such as ChatGPT. Two reasons rule it out. First, \emph{data sovereignty}: educational records contain highly sensitive personally identifiable information, and routing them to foreign servers directly violates national data-localization laws such as Vietnam's Decree 53 \cite{vietnam_decree53}, compelling institutions to self-host open models strictly on-premise. Second, \emph{curriculum grounding}: pre-trained on English-dominant, Western-centric corpora \cite{cahyawijaya2024llms}, such models are not organized around the Vietnamese textbook curriculum (\emph{Sach Giao Khoa}, SGK); their knowledge of local content is unsystematic and frequently hallucinated \cite{vnexpress_chatgpt_education_2023}.\footnote{For instance, early deployments of ChatGPT in Vietnam generated severe historical hallucinations, such as claiming national leaders were founders of the World Health Organization (WHO), or confidently asserting that Emperor Quang Trung and his birth name Nguyen Hue were two distinct kings from opposing dynasties \cite{dnto_chatgpt_2023}. Such algorithmic hallucinations highlight the severe pedagogical risks of deploying ungrounded models in regional education.} A trustworthy tutor must therefore be both self-hosted and explicitly grounded in localized curricular knowledge.

Both requirements are hard on the infrastructure available to public schools. Self-hosting on limited hardware creates a two-fold bottleneck. The static footprint of model weights can be tamed by post-training quantization such as AWQ \cite{lin2024awq} or GPTQ \cite{frantar2023gptq}; the \emph{dynamic} cost of long contexts cannot. A tutoring session returns to the same material turn after turn---textbook chapters, exercises and the dialogue so far---so its prompts share long prefixes, and serving engines reuse their KV Cache instead of recomputing it \cite{kwon2023efficient, zheng2024sglang}. Such reuse is the norm in production: Character.AI reports a $95\%$ cache hit rate on its dialogues \cite{characterai2024inference}, and cloud and agent workloads show the same pattern \cite{wang2025kvcachewild, ji2025manus}. Prefill is therefore largely paid once, and decoding dominates what the student waits for: every generated token reads the whole KV Cache, so decoding slows down as the session grows and as more students share the same GPU---a bottleneck that I/O-aware kernels such as FlashAttention \cite{dao2022flashattention} and PagedAttention \cite{kwon2023efficient} alleviate but do not remove. Sparse decoders reduce this cost, but they either re-score the cache at every decoding step \cite{tang2024quest, omnikv2025} or evict entries for good \cite{li2024snapkv, feng2024adakv}, losing what a later question in the same session may need. Grounding is equally costly the obvious way: continually fine-tuning on evolving curricula is computationally prohibitive and risks catastrophic forgetting \cite{efficientcpt2025naacl}. Context adaptation avoids weight updates by accumulating a playbook of lessons from past interactions \cite{agenticcontext2026}, yet a growing playbook does not by itself generalize better.

We address both obstacles with \textbf{DeepEdu-F}, a framework for DeepEdu. Its inference engine, \textbf{LastSet}, selects the KV Cache entries a question needs once, decodes the whole turn from that set, and keeps the full cache for later turns. Its agentic layer, \textbf{REPAIR} (Reliable Evidence-based Playbook Adaptation and Insight Revision), curates a verified playbook from past interactions. The main contributions of our work are as follows:

\begin{enumerate}
    \item We introduce \textit{LastSet}, a long-context inference engine that selects the critical KV Cache entries once per turn from the attention of the question and shares the selected set across all layers and decoding steps. It decodes up to $3.1\times$ faster than dense decoding and faster than four sparse decoding baselines, with accuracy close to full attention.

    \item We propose \textit{REPAIR}, a self-improving agentic layer that curates a verified playbook without weight updates: its Curator revises, merges and deletes entries as well as adding them, and its Reflector sees earlier failures confirmed against the reference.
    \item We evaluate REPAIR on financial-reasoning and interactive-agent benchmarks against a strong context-adaptation baseline.

\end{enumerate}

\section{Related Work}
\label{sec:related}
To contextualize the contributions of DeepEdu-F, this section surveys
recent advancements across the architectural evolution of LLMs and 
their application to education, the principal techniques for 
compressing models and optimizing long-context inference under hardware 
constraints, the challenges of deploying LLMs on low-resource languages 
and within localized cultural contexts, and the emerging paradigm of 
self-improving agentic frameworks.

\subsection{The Evolution of Large Language Models}
The landscape of Natural Language Processing (NLP) has been fundamentally transformed by the scaling of neural architectures, passing through several critical milestones. This trajectory is rooted in the Transformer architecture and early scaling laws \cite{vaswani2017attention, brown2020language}. Subsequently, the community witnessed the rise of powerful open-weights foundation models that democratized high-performance language modeling \cite{touvron2023llama, bai2023qwen}. To optimize the intrinsic trade-off between model capacity and inference cost, recent architectures have increasingly adopted the Mixture-of-Experts (MoE) paradigm \cite{yang2024qwen2, qwen252024, jiang2024mixtral}. This architectural evolution has recently culminated in highly efficient, post-training reinforced paradigms (such as DeepSeek-R1 and Qwen3) that emphasize advanced logical reasoning and test-time scaling over pure pattern matching \cite{deepseek2024v3, qwen32025}.

Concurrently, a pivotal paradigm shift has occurred: moving from static sequence-to-sequence generation toward interactive, tool-augmented autonomous agents. This transition was initially catalyzed by models pre-trained on vast codebases, such as Codex \cite{chen2021codex}, which demonstrated that LLMs could translate natural language into actionable programmatic commands. Building upon this, frameworks like ReAct \cite{yao2023react} and Toolformer \cite{schick2023toolformer} equipped models with the ability to synergize step-by-step reasoning with external API invocations. As reasoning engines matured, the focus shifted toward multi-agent collaboration and fully autonomous software engineering capabilities. Recent advancements, ranging from open-source frameworks like AutoGen and SWE-agent \cite{wu2023autogen, yang2024sweagent} to highly integrated proprietary systems like Claude Code \cite{anthropic2025claudecode}, showcase agents capable of navigating complex file systems, executing code iteratively, and autonomously self-correcting. Collectively, these advances established LLM agents as a viable substrate for complex, interactive applications — among which personalized education has emerged as a particularly compelling domain.

\subsection{Foundation Models for Education}
The integration of generative AI into educational technology represents a profound shift from static, rule-based Intelligent Tutoring Systems (ITS) to dynamic, generative frameworks. Early approaches primarily relied on rigid, operation-based formalisms to parse and solve domain-specific tasks, such as mathematical word problems \cite{amini2019mathqa}. However, the advent of Large Language Models has fundamentally upgraded this paradigm. Comprehensive surveys now outline the transition toward utilizing foundation models as core cognitive engines for personalized education, emphasizing their capacity to dynamically tailor feedback, simulate roleplay, and adapt to individual learning styles \cite{llmeducation2026}.

To bridge the gap between general-purpose text generation and strict pedagogical utility, recent advancements have focused on developing specialized, agentic tutoring frameworks. For instance, models such as MathCoder \cite{wang2024mathcoder} interleave natural language reasoning with verifiable code execution, ensuring that students receive accurate, step-by-step guidance rather than hallucinated solutions. Furthermore, the integration of LLMs with Knowledge Tracing (KT) paradigms enables these agents to actively model a student's evolving cognitive state. By maintaining dynamic representations of a learner's historical performance, state-of-the-art educational agents can autonomously adapt their pedagogical scaffolding—seamlessly shifting between direct hints and deeper Socratic questioning based on real-time comprehension \cite{macina2023mathdial}. Nevertheless, the practical deployment of such educational agents at scale remains constrained by two orthogonal challenges that have received comparatively less attention: the computational cost of running these models on accessible hardware, and their tendency to misalign with localized curricula — issues we now examine in turn.

\subsection{Post-Training Quantization for Efficient Deployment}
\label{subsec:quantization}

A fundamental prerequisite for deploying state-of-the-art LLMs on consumer-grade hardware is reducing the static memory footprint of model weights. Post-training quantization (PTQ) has emerged as the dominant solution, compressing pre-trained weights into low-bit representations without expensive re-training. GPTQ~\cite{frantar2023gptq} pioneered accurate one-shot quantization for generative transformers, employing an approximate second-order procedure to minimize layer-wise reconstruction error at 3 or 4 bits. AWQ~\cite{lin2024awq} further observes that weight importance is highly non-uniform: a small fraction of salient weights, identified through activation magnitudes, disproportionately influences output quality, and selectively preserving them throughper-channel scaling achieves superior accuracy under hardware-friendly uniform bit-widths. However, quantization addresses only the static weight footprint; the dynamic memory growth of the KV Cache during long-context inference remains entirely unresolved, motivating the complementary techniques surveyed next.

\subsection{Knowledge Distillation for Compact Educational Models}
\label{subsec:distillation}

An alternative compression paradigm transfers the capabilities of large 
teacher models into smaller student architectures through knowledge 
distillation (KD). Traditional KD~\cite{hinton2015distilling} matches 
soft probability distributions, but often induces hallucinations in 
generative tasks due to the mode-covering behavior of forward 
Kullback-Leibler divergence. MiniLLM~\cite{gu2024minillm} mitigates 
this through reverse KLD with policy gradient optimization, forcing 
students to concentrate probability mass on high-quality generation 
regions. Beyond probability matching, Hsieh et al.~\cite{hsieh2023distilling} 
introduced step-by-step distillation that extracts intermediate reasoning 
trajectories from proprietary teachers, enabling compact students to 
internalize multi-step problem-solving heuristics --- a capability 
particularly valuable for pedagogical scaffolding in educational tutoring.

Nevertheless, distillation pipelines remain computationally intensive 
at training time and produce student models with static knowledge that 
cannot adapt to evolving curricula without repeated re-distillation, 
limiting their applicability for institutions facing both hardware and 
content-update constraints.

\subsection{Long-Context Inference Optimization}
\label{subsec:long-context}
Positional interpolation, long-context post-training and memory modules
extend the context a model can read~\cite{chen2023extending,
peng2024yarn, dai2019transformerxl}; we take such a model as given and
target the cost of serving it. Serving engines keep the KV Cache of
processed prompts and reuse it for later requests that share a
prefix~\cite{kwon2023efficient, zheng2024sglang}, and production traces
show that such reuse is frequent~\cite{characterai2024inference,
wang2025kvcachewild}. Sparse prefill methods such as
MInference~\cite{jiang2024minference} accelerate the first pass over a
long prompt, but with prefix caching most of a session is prefilled once,
so its cost moves mostly to decoding. Each decode step reads the whole KV Cache,
so decoding is bound by memory bandwidth, which FlashAttention and
PagedAttention use efficiently but do not reduce~\cite{dao2022flashattention,
kwon2023efficient}.

Sparse decoding reads only part of the cache, and existing methods fall
into three families. \emph{Eviction} methods keep a fixed subset and
discard the rest, chosen as attention sinks plus a recent
window~\cite{xiao2024efficient, han2024lminfinite}, by accumulated
attention~\cite{zhang2024h2o}, or from the attention of the prompt's last
queries, as in SnapKV~\cite{li2024snapkv} and AdaKV~\cite{feng2024adakv};
they decode fast but lose entries that a later question may need.
\emph{Per-step selection} methods keep the full cache and choose entries
again at every decoding step, over memory blocks~\cite{xiao2024infllm},
pages~\cite{tang2024quest}, a few important channels~\cite{sparseattention2024}
or a low-rank key cache~\cite{sun2025shadowkv}; they preserve later turns
but pay a scoring cost at every step. \emph{Cross-layer reuse} reduces
that cost by selecting at a few layers and letting the others attend to
the same tokens, as in OmniKV~\cite{omnikv2025} and
TidalDecode~\cite{yang2024tidaldecode}, but still selects again at every
step. A decoder that selects once per turn, shares that selection across
layers and decoding steps, and keeps the full cache for later turns lies
between these families.

\subsection{LLMs on Low-Resource Languages}
While modern LLMs exhibit remarkable zero-shot and few-shot reasoning capabilities in high-resource languages like English, their performance often degrades significantly when applied to low-resource languages (LRLs) such as Vietnamese. This degradation is primarily attributed to a severe underrepresentation of LRL tokens in the pre-training corpora, which limits the models' native linguistic alignment. 

To circumvent this linguistic bottleneck without resorting to computationally prohibitive pre-training from scratch, recent literature has explored several efficient adaptation strategies. At the inference level, Nguyen et al. \cite{democratizing2024acl} propose leveraging the models' dominant English capabilities through linguistically-diverse prompts. This approach acts as a cross-lingual bridge, effectively anchoring LRL inputs to the models' highly structured English latent space to improve zero-shot generalization. Complementarily, Cahyawijaya et al. \cite{cahyawijaya2024llms} demonstrate that LLMs possess latent cross-lingual capabilities that can be explicitly activated via In-Context Learning (ICL). Their empirical findings show that providing a limited number of task-specific demonstrations in the target language allows the model to dynamically adapt its reasoning pathways to the LRL without requiring any parameter updates. 

For applications requiring deeper structural alignment, prompt engineering alone is often insufficient. To address this, Nag et al. \cite{efficientcpt2025naacl} introduce efficient continual pre-training (CPT) techniques tailored for LRLs. Their methodology natively injects new linguistic knowledge into the foundation model's weights while actively mitigating the catastrophic forgetting of general reasoning capabilities, all under strict computational constraints. Collectively, these methodologies highlight that effective LRL 
adaptation can be achieved without invasive parameter updates, 
particularly when combined with carefully curated in-context signals.

\subsection{Hallucination in Localized and Cultural Contexts}
While modern LLMs demonstrate strong generalization capabilities, their deployment in region-specific applications is frequently hindered by factual hallucinations. Ji et al. \cite{ji2023survey} identify that hallucinations fundamentally stem from data divergence, where models learn spurious correlations from imbalanced pre-training corpora. This issue is severely amplified by the Western-centric nature of foundational models. As Huang et al. \cite{huang2023survey} highlight, when LLMs process queries outside their dominant data distribution, they often generate confident confabulations rather than acknowledging knowledge boundaries. Empirical evaluations by Bang et al. \cite{bang2023multitask} confirm this limitation, demonstrating that LLMs exhibit significantly higher hallucination rates and degraded reasoning when applied to low-resource languages and non-Western cultural contexts. In educational settings, relying on such ungrounded models introduces critical pedagogical risks. These findings establish that scaling pre-training alone cannot resolve 
the localization gap, particularly in educational deployments where 
factual accuracy on regional content is non-negotiable. This persistent 
gap motivates runtime grounding mechanisms capable of injecting 
localized knowledge without the prohibitive cost of continual 
re-training.

\subsection{Agentic Systems and Self-Evolving Context}
\label{subsec:agentic}
The rapid evolution of LLMs has introduced a paradox in knowledge 
management: while models store vast generalized knowledge within their 
parametric memory, their ability to seamlessly integrate deeply 
specialized domain heuristics remains limited. Traditional fine-tuning 
is rigid, computationally expensive, and susceptible to catastrophic 
forgetting. Conversely, current prompt optimization methods often 
suffer from brevity bias --- the tendency to compress prompts 
into short, generic instructions that strip away essential technical 
nuances required for high-precision tasks~\cite{promptalchemist2025}. 
Furthermore, attempts at monolithic rewriting in long-context windows 
frequently trigger context collapse, wherein iterative LLM 
rewrites progressively erode accumulated domain knowledge into 
uninformative summaries.

To overcome these barriers and build agentic systems that self-improve without weight updates, recent literature explores natural language feedback and verbal reinforcement learning \cite{krause2019dynamic, shinn2023reflexion, nature2025optimizing}. Techniques leveraging reflective prompt evolution \cite{gepa2025} and agentic context engineering \cite{agenticcontext2026} effectively transform the static context into a living, continuously updated "playbook." These methodologies enable LLM agents to adapt and fine-tune their behavior exclusively through dynamic context memory, preserving granular domain insights without the massive overhead of standard fine-tuning~\cite{memento_no_date}.

\subsection{Discussion and Positioning}
\label{subsec:rw-positioning}

Table~\ref{tab:positioning} summarizes where DeepEdu-F sits relative to the
four directions above. Each addresses one facet of on-premise educational
deployment but leaves the others open: quantization shrinks static weights
yet does nothing for the dynamic long-context cost; selective sparse
attention accelerates long contexts but neither self-improves nor
localizes; agentic context engineering self-improves without weight
updates but is not built for long-context efficiency or a specific
curriculum; and low-resource-language adaptation localizes, but through
weight updates that are costly to maintain as curricula evolve. DeepEdu-F is,
to our knowledge, the first framework to occupy their intersection: it
couples a turn-level sparse decoding engine with a self-improving,
curriculum-grounded agentic layer, so that a single on-premise system is
efficient, self-improving, and localizable at once.

\begin{table}[t]
\centering
\caption{Positioning of DeepEdu-F against representative prior directions.
\checkmark: directly addressed; (\checkmark): partial; --: not addressed.
Columns: on-premise/data-sovereignty fit, long-context efficiency,
self-improvement without fine-tuning, and localization.}
\label{tab:positioning}
\footnotesize
\setlength{\tabcolsep}{3pt}
\begin{tabular}{lcccc}
\toprule
\textbf{Direction} & \textbf{On-prem.} & \textbf{Long-ctx} & \textbf{Self-impr.} & \textbf{Local.} \\
\midrule
Quantization~\cite{lin2024awq, frantar2023gptq}                  & \checkmark & -- & -- & -- \\
Sparse long-context~\cite{tang2024quest, omnikv2025}       & -- & \checkmark & -- & -- \\
Agentic context~\cite{agenticcontext2026, shinn2023reflexion}    & (\checkmark) & -- & \checkmark & -- \\
LRL adaptation~\cite{efficientcpt2025naacl, cahyawijaya2024llms} & -- & -- & -- & \checkmark \\
\midrule
\textbf{DeepEdu-F (ours)}                                        & \checkmark & \checkmark & \checkmark & (\checkmark) \\
\bottomrule
\end{tabular}
\end{table}

\section{Preliminaries}
\label{sec:preliminaries}

This section establishes the notation and the two paradigms our framework
builds on. We first summarize Transformer inference and Multi-Head
Attention (§\ref{subsec:llm-inference}) and formalize long-context
inference as a Selective Sparse Attention problem
(§\ref{subsec:problem-formulation})---the basis of our inference
engine---then define Agentic Context Engineering
(§\ref{subsec:prelim-ace}), the paradigm underlying our self-improving
agentic layer.

\subsection{LLM Inference and Multi-Head Attention}
\label{subsec:llm-inference}

\paragraph{Notation}
Throughout this paper, let $d$ denote the model hidden dimension, $H$ the
number of query heads, and $d_h = d/H$ the per-head dimension. Under
grouped-query attention (GQA)~\cite{ainslie2023gqa}, the query heads are
partitioned into $G$ groups $H_1, \ldots, H_G$, and the heads of group $g$
share one key/value head. A layer processes the $n$ tokens of the current
pass, and the KV Cache accumulated from previously processed tokens has
length $N$.

\paragraph{Multi-head attention}
Given hidden states $\mathbf{X} \in \mathbb{R}^{n \times d}$, each query
head $h \in H_g$ computes queries $\mathbf{Q}^{(h)} \in \mathbb{R}^{n \times d_h}$,
and group $g$ computes keys and values
$\mathbf{K}^{(g)}, \mathbf{V}^{(g)} \in \mathbb{R}^{n \times d_h}$. Each head
applies scaled dot-product attention (SDPA)~\cite{vaswani2017attention}:
\begin{equation}
    \mathbf{O}^{(h)} = \text{softmax}\!\left(\frac{\mathbf{Q}^{(h)} (\mathbf{K}^{(g)})^\top}{\sqrt{d_h}}\right) \mathbf{V}^{(g)},
    \label{eq:sdpa}
\end{equation}
and the per-head outputs are concatenated and projected back to
dimension $d$.

\paragraph{Prefill and decode}
Inference has two stages. \emph{Prefill} processes the $n$ prompt tokens
of a turn in one pass, at a cost of $\mathcal{O}(n^2 d_h)$ per head, and
stores their keys and values in the KV Cache
$\mathbf{K}^{(g)}_{\text{cache}}, \mathbf{V}^{(g)}_{\text{cache}} \in \mathbb{R}^{N \times d_h}$;
with prefix caching, a later turn prefills only its new tokens. \emph{Decode}
then generates one token per pass ($n = 1$) and appends its key and value,
so $N$ grows by one per step. Each decode step reads all $N$ cached entries
of every layer, $\mathcal{O}(N d_h)$ per head, so decoding is bound by
memory bandwidth and slows down as the context grows and as concurrent
requests share the GPU. Reducing the number of entries each step reads is
the target of our inference engine.

\subsection{Selective Sparse Attention}
\label{subsec:problem-formulation}

The cost of reading the whole KV Cache at every decode step, combined with
the abundant sparsity empirically observed in LLM attention
maps~\cite{xiao2024infllm, tang2024quest}, motivates
the selective sparse attention (SSA) paradigm: rather than attending to
the entire KV Cache, the current queries attend only to a small subset of
critical entries. Our inference engine builds directly on this paradigm;
we restate it below in our notation.

\begin{definition}[Selective Sparse Attention]
\label{def:ssa}
Consider a transformer layer at any inference step, with queries
$\mathbf{Q}^{(h)} \in \mathbb{R}^{n \times d_h}$ for the $n$ current tokens
and, for each KV group $g$, a KV Cache
$\mathbf{K}^{(g)}_{\text{cache}}, \mathbf{V}^{(g)}_{\text{cache}} \in \mathbb{R}^{N \times d_h}$.
For a head $h \in H_g$, the full attention output $\mathbf{O}^{(h)}$ is
Equation~\eqref{eq:sdpa} with $\mathbf{K}^{(g)}_{\text{cache}}$ and
$\mathbf{V}^{(g)}_{\text{cache}}$ as keys and values. Selective sparse
attention restricts this softmax to a subset of the cache entries,
yielding the sparse approximation $\hat{\mathbf{O}}^{(h)}$:
\begin{equation}
    \hat{\mathbf{O}}^{(h)} = \text{softmax}\!\left(\frac{\mathbf{Q}^{(h)} (\mathbf{K}^{(g)}_{\text{select}})^\top}{\sqrt{d_h}}\right) \mathbf{V}^{(g)}_{\text{select}}.
    \label{eq:sparse-attention}
\end{equation}
The two formulations differ only in \emph{which} cache rows enter the
softmax: $\mathbf{K}^{(g)}_{\text{select}}$ and
$\mathbf{V}^{(g)}_{\text{select}}$ are the rows of
$\mathbf{K}^{(g)}_{\text{cache}}$ and $\mathbf{V}^{(g)}_{\text{cache}}$ at
the positions $\mathcal{I}_g$ chosen by a \textit{selection function}
$\mathcal{S}$ from the queries of the group:
\begin{equation}
    \mathcal{I}_g = \mathcal{S}\!\left(\{\mathbf{Q}^{(h)}\}_{h \in H_g},\, \mathbf{K}^{(g)}_{\text{cache}}\right),
    \label{eq:selection}
\end{equation}
with $\mathcal{I}_g \subseteq \{1, \ldots, N\}$ and $|\mathcal{I}_g| \ll N$.
All query heads of a group thus read the same selected entries, matching
the shared key/value head of GQA. The design objective is to construct
$\mathcal{S}$ such that the approximation gap
$\sum_{h \in H_g} \big\|\mathbf{O}^{(h)} - \hat{\mathbf{O}}^{(h)}\big\|_2^2$
stays small while $|\mathcal{I}_g|$ fits the memory-read budget.
\end{definition}

Beyond \emph{what} they select (§\ref{subsec:long-context}), selection
functions differ in \emph{how often} $\mathcal{S}$ runs: at every decode
step, conditioned on the current query; once at prefill, discarding the
entries outside $\mathcal{I}_g$; or, as in our engine, once per turn while
keeping the full cache (§\ref{subsec:long-context-engine}).

\subsection{Agentic Context Engineering}
\label{subsec:prelim-ace}

While selective sparse attention addresses the computational bottleneck,
the semantic bottleneck---grounding the model in localized, trustworthy
knowledge---calls for a complementary paradigm. Our agentic layer builds
on Agentic Context Engineering (ACE)~\cite{agenticcontext2026}, which
adapts an LLM agent by editing its context rather than its weights. We
restate it in the notation used later.

\begin{definition}[Agentic Context Engineering]
\label{def:ace}
Let a backbone LLM with fixed parameters $\phi$ process a stream of
interactions $\mathcal{D}=\{(q_t,c_t,y_t)\}_{t=1}^{T}$, where $q_t$ is a
query, $c_t$ its available context, and $y_t$ a feedback signal (e.g., a
reference answer or a deterministic environment outcome). ACE maintains a
structured \emph{playbook} $\mathcal{P}_t$ of itemized natural-language
entries and, after each interaction, applies a set of edits
$\Delta_t$ distilled from the resulting trajectory:
\begin{equation}
    \mathcal{P}_{t+1}=\operatorname{Apply}(\mathcal{P}_t,\Delta_t),
    \qquad \phi_{t+1}=\phi_t.
    \label{eq:ace-context-update}
\end{equation}
Adaptation therefore occurs entirely in inspectable context, leaving the
weights $\phi$ unchanged.
\end{definition}

Equation~\eqref{eq:ace-context-update} leaves open two design choices that
decide whether such an agent actually improves: \emph{how} the edits
$\Delta_t$ are produced, and \emph{which} entries condition each
interaction as the playbook grows. Our agentic layer, REPAIR, builds on the Generator--Reflector--Curator
loop of ACE and resolves both choices: Section~\ref{sec:motivations}
shows how they fail in practice, and §\ref{subsec:agentic-engine}
presents REPAIR.

\section{Motivations and Observations}
\label{sec:motivations}

The design of DeepEdu-F follows from four observations: two on decoding
cost, which motivate LastSet (§\ref{subsec:obs-decode}--\ref{subsec:obs-tradeoff}),
and two on playbook adaptation, which motivate the two parts of REPAIR
(§\ref{subsec:obs-curation}--\ref{subsec:obs-agentic}).

\subsection{Decoding Dominates Later Turns}
\label{subsec:obs-decode}

\begin{figure}[t]
\centering
\includegraphics[width=\columnwidth]{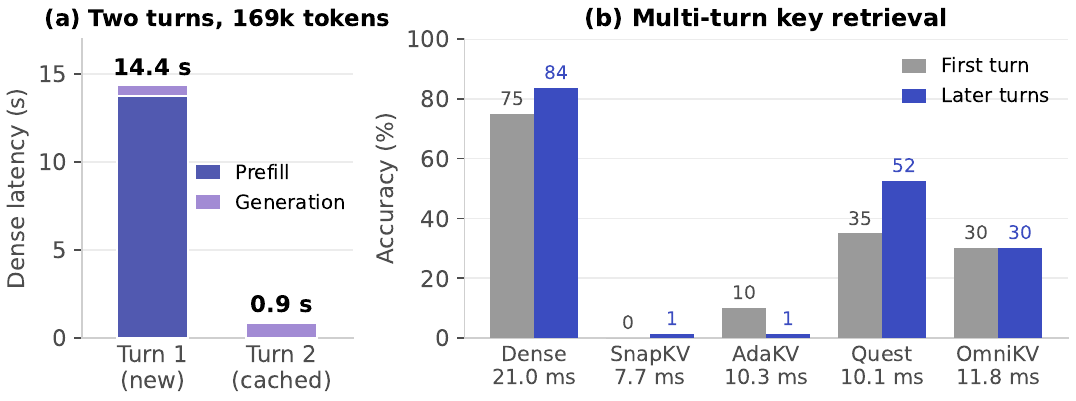}
\caption{Multi-turn sessions with Qwen2.5-7B-Instruct-1M on one H100.
(a) Dense latency of two questions on the same $169$k-token context; the
second turn reuses the cached prefix. (b) Key retrieval in the multi-turn
sessions of SCBench at budget $k = 8{,}192$; the time per output token at
$128$k with six requests is given under each method.}
\label{fig:motiv}
\end{figure}

With prefix caching, a later question on the same context prefills only
its new tokens: on a $169$k-token context, the second question takes
$0.9$\,s against $14.4$\,s for the first, and most of it is spent
generating the answer (Figure~\ref{fig:motiv}a). Each decode step,
however, reads the whole KV Cache, so decoding slows down as the context
grows and as concurrent requests share the memory bandwidth: at $128$k
tokens, the time per output token rises from $8.5$\,ms with one request
to $21.0$\,ms with six (Table~\ref{tab:lastset-tpot}). Once
the context is cached, the number of entries each decode step reads
determines response time.

\subsection{Sparse Decoders Trade Speed for Exact Retrieval}
\label{subsec:obs-tradeoff}

Figure~\ref{fig:motiv}b runs the multi-turn key-retrieval sessions of
SCBench~\cite{li2025scbench}, in which several questions share one long
context. The eviction methods SnapKV and AdaKV keep only the entries chosen at
prefill and score close to zero, although SnapKV is the fastest
decoder. Quest and OmniKV keep the full cache and choose new entries at every
step. On later turns they reach $52\%$ and $30\%$ accuracy, against
$84\%$ for dense decoding, and choosing at every step makes them
slower than SnapKV. No existing decoder is both fast and exact
across turns, which motivates selecting less often while keeping the full
cache (§\ref{subsec:long-context-engine}).

\subsection{Playbook Growth Does Not Bring Better Generalization}
\label{subsec:obs-curation}

\begin{figure}[!t]
\centering
\includegraphics[width=\columnwidth]{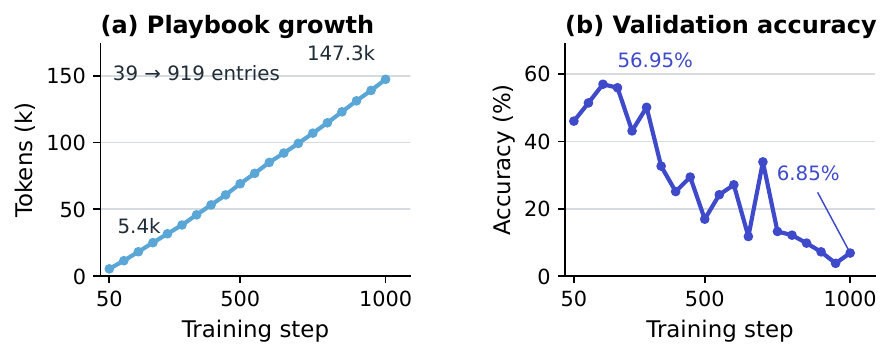}
\caption{An add-only Curator on FiNER. (a) Playbook size during
training. (b) Per-label validation accuracy.}
\label{fig:obs-curation-growth}
\end{figure}

ACE adapts an agent by adding lessons to its
Playbook~\cite{agenticcontext2026}. We follow an ACE agent whose Curator
can only add entries through $1{,}000$ training steps on FiNER
(Figure~\ref{fig:obs-curation-growth}). The Playbook grows steadily to
$919$ entries and about $147$k tokens, whereas validation accuracy peaks
at $56.95\%$ after $150$ steps and then falls to $6.85\%$.
Checking the validation outputs confirms the endpoint: only $137$ of
$2{,}000$ labels are correct. The agent also struggles to return the
required answer: samples with no extractable final answer rise from
$88/500$ ($17.6\%$) at step $150$ to $359/500$ ($71.8\%$) at step
$1{,}000$, while logged errors with an extractable answer but the wrong
number of labels rise from $9$ to $94$. At least $90.6\%$ of the final
validation samples therefore lack an extractable answer or fail to
return the required four labels.

More lessons therefore do not mean better guidance on unseen tasks:
the Playbook keeps growing, but the quality of the agent's answers
falls sharply. This pattern suggests \emph{context pollution}, where
irrelevant or competing rules can accumulate alongside useful ones
and make the task instructions harder to follow. Once an unhelpful
entry enters an add-only Playbook, it stays there, because new entries
are only added next to it and never correct or remove it. A Curator
should therefore also be able to revise, merge and delete entries
(\S\ref{subsubsec:ace-curator}).

\subsection{Systematic and Recurring Agent Failures}
\label{subsec:obs-agentic}

Adding a lesson after a failure does not necessarily prevent a related
mistake on a later task. We examine this recurrence in a $500$-task
offline Formula run, which produces $142$ failures verified against
the ground truth ($28.4\%$ of tasks). Two failure observations are
considered semantically related when the cosine similarity between
BGE-M3 embeddings of their concatenated error descriptions and
root-cause diagnoses exceeds $0.8$; we use
\texttt{BAAI/bge-m3}.\footnote{\url{https://huggingface.co/BAAI/bge-m3}}
For each failure, we count only related failures that occur later
in the training sequence.

\begin{figure}[!htbp]
\centering
\includegraphics[width=0.96\columnwidth]{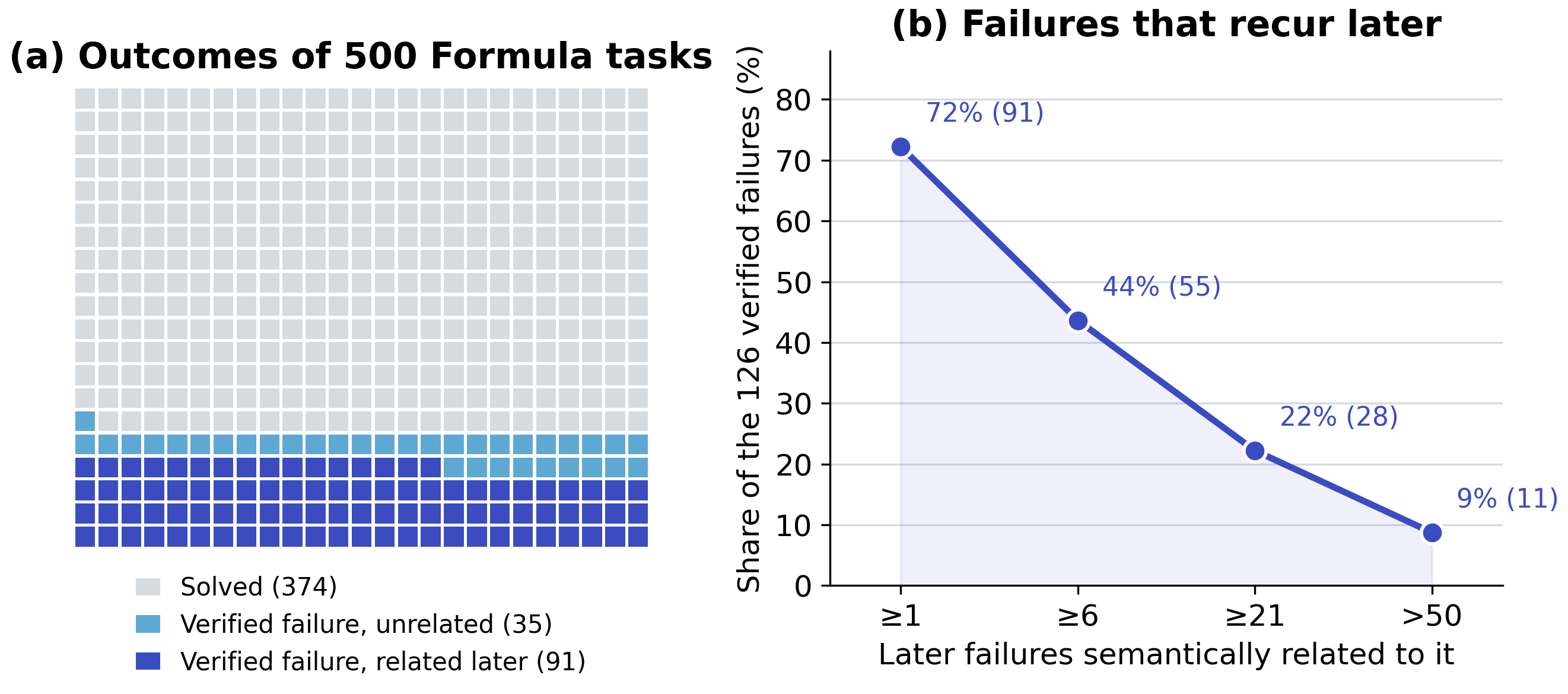}
\caption{Failures in the $500$-task offline run on Formula. (a) Outcome of
each task: solved, or a failure verified against the ground truth that is
or is not semantically related to a later failure. (b) Share of the
verified failures related to at least the given number of later failures.}
\label{fig:noise}
\end{figure}

Figure~\ref{fig:noise} shows that $115$ of the $142$ verified failures
($81.0\%$) are related to at least one later failure; $73$ ($51.4\%$)
are related to at least five, and $17$ ($12.0\%$) to at least twenty.
The remaining $27$ have no later match under this criterion.
Thus, failures are not merely isolated events: related error patterns
reappear across subsequent tasks despite the intervening adaptation.

The observation complements \S\ref{subsec:obs-curation}: maintaining
the Playbook is necessary, but a new diagnosis should also be informed
by what went wrong before. If each failure is analyzed in isolation,
the agent can repeatedly rediscover a weakness without connecting it
to earlier evidence or attempted corrections. Retaining verified
failures and making analogous diagnoses available to the Reflector
therefore motivates the failure-memory grounding in
\S\ref{subsubsec:ace-fmb}.

\section{Methodology}
\label{sec:methodology}

This section presents the two components of DeepEdu-F.
§\ref{subsec:long-context-engine} introduces LastSet, which addresses the
decoding cost of §\ref{subsec:obs-decode}--\ref{subsec:obs-tradeoff}: it
selects the critical KV Cache entries (Definition~\ref{def:ssa}) once per
turn, decodes the whole turn from that set and keeps the full cache for
later turns. §\ref{subsec:agentic-engine} introduces REPAIR, which extends
ACE (Definition~\ref{def:ace}) with one part for each observation in
§\ref{subsec:obs-curation}--\ref{subsec:obs-agentic}. REPAIR decides what
the context of the agent contains, and LastSet makes that long context
fast to answer.

\subsection{Efficient Long-Context Inference Engine}
\label{subsec:long-context-engine}

This subsection presents LastSet, a sparse decoding engine that invokes
the selection function $\mathcal{S}$ of Equation~\eqref{eq:selection}
\emph{once per turn}, at the end of prefill, instead of at every decode
step. Every layer and every decode step of the turn then reads only the
selected entries, while the full KV Cache stays in memory so that the next
turn can select again.

\begin{figure*}[t]
    \centering
    \includegraphics[width=0.95\textwidth]{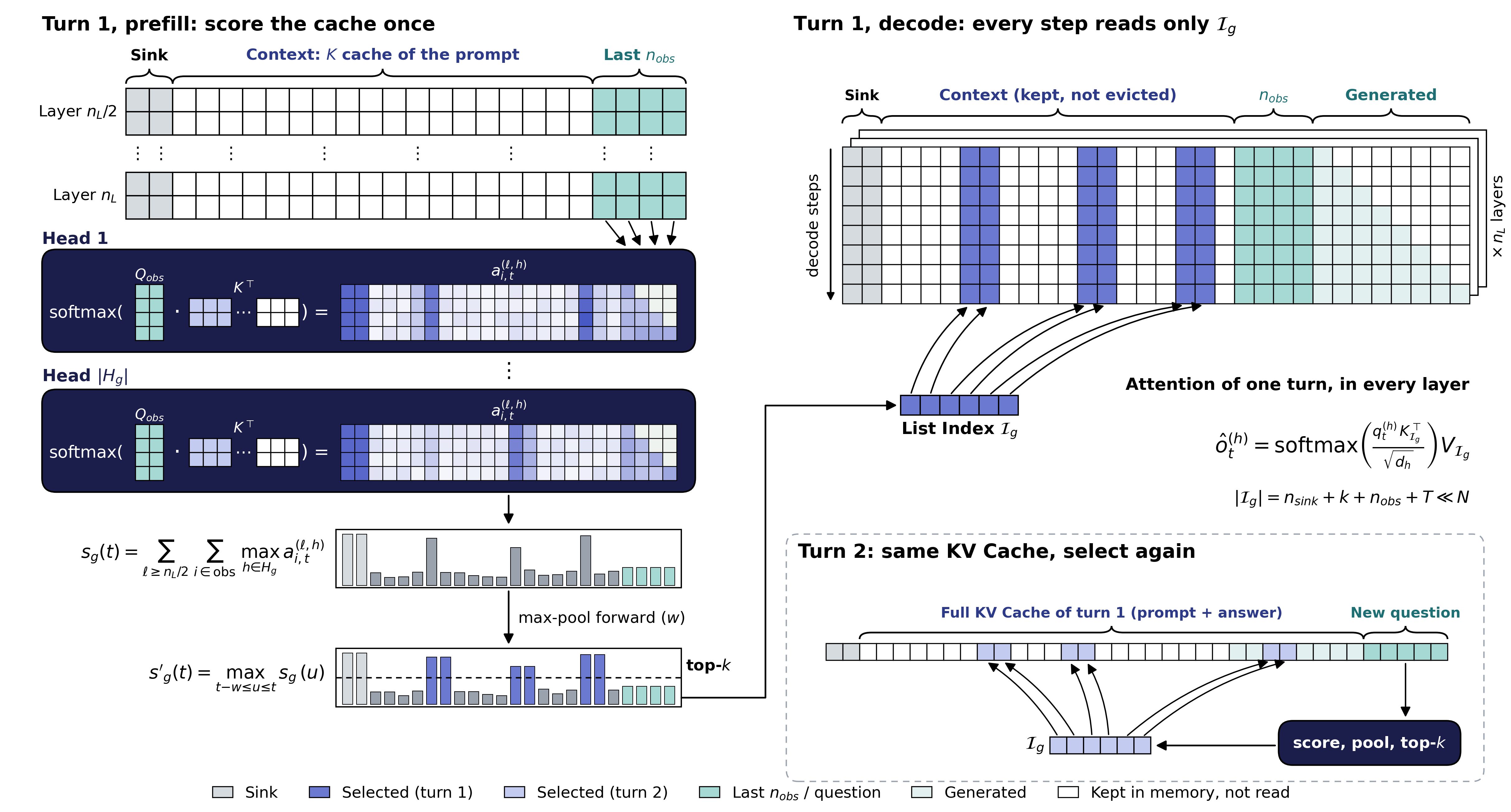}
    \caption{LastSet. Turn 1: at the end of prefill, the attention of the last
    $n_{\text{obs}}$ prompt queries, summed over the upper half of the
    layers and max-pooled forward, scores every cached token; LastSet keeps
    the attention sinks, the top-$k$ tokens and a recent window, one set
    $\mathcal{I}_g$ per KV group, and every layer and decode step of the turn
    reads only $\mathcal{I}_g$. The full KV Cache stays in memory. Turn 2:
    selection runs again over the full cache, so the new question can reach
    any token.}
    \label{fig:lastset-method}
\end{figure*}

Section~\ref{sec:motivations} sets three requirements for such a decoder.
Each decode step should read only a small part of the cache, since its
cost grows with the number of entries it reads
(§\ref{subsec:obs-decode}); choosing these entries should not add work to
every step, which is what slows down Quest and OmniKV; and no entry may be
discarded for good, which is what makes SnapKV and AdaKV fail on later
turns (§\ref{subsec:obs-tradeoff}). One selection per turn over a cache
that is never pruned meets all three, provided that the entries chosen
before the first token is generated serve the whole answer. The rest of
this subsection describes how LastSet scores the cache
(§\ref{subsubsec:lastset-score}), which entries it keeps and how decoding
uses them (§\ref{subsubsec:lastset-set}), how it handles the later turns
of a session (§\ref{subsubsec:lastset-turns}), and what it costs
(§\ref{subsubsec:lastset-cost}). Figure~\ref{fig:lastset-method}
illustrates the procedure.

\subsubsection{Turn-level token scoring}
\label{subsubsec:lastset-score}

Let $n_L$ be the number of layers, $n$ the prompt length of the current
turn, and $a^{(\ell,h)}_{i,t}$ the attention weight of query $i$ on cached
token $t$ in layer $\ell$ and head $h$, computed during prefill. The last
$n_{\text{obs}}$ prompt tokens hold the student's question and the chat
template, so their queries indicate what the answer will need; they are
also the only queries available before decoding starts. Under
grouped-query attention, the query heads $H_g$ of KV group $g$ share one
key/value head; LastSet scores each cached token for each KV group as
\begin{equation}
    s_g(t) = \sum_{\ell \geq n_L/2} \; \sum_{i = n - n_{\text{obs}} + 1}^{n} \; \max_{h \in H_g} \, a^{(\ell,h)}_{i,t}.
    \label{eq:lastset-score}
\end{equation}
Two choices in Equation~\eqref{eq:lastset-score} follow from how
attention behaves. First, the score sums only the upper half of the
layers, because the tokens they rank highest are stable from layer to
layer. Following ASL~\cite{asl2026}, we measure the variance of the rank of
the top tokens over windows of four consecutive layers: on four long-context
retrieval tasks, it is $16$--$45\%$ lower in the upper half than in the
lower half, so the upper layers agree on which tokens matter. Second, the
score takes the maximum over the query heads of a group rather than their
mean. The evidence for a question is often found by a few retrieval
heads, fewer than $5\%$ of all heads~\cite{wu2025retrieval}; under a mean
over the $|H_g|$ heads of a group, a token that only one head needs would
lose to tokens that every head attends to weakly, whereas the maximum keeps
it.

The attention weights of Equation~\eqref{eq:lastset-score} are not
returned by the prefill kernel, since FlashAttention never materializes
the attention matrix. In each layer of the upper half, right after the
prefill attention of that layer, LastSet therefore recomputes the
attention of the last $n_{\text{obs}}$ queries against all $N$ cached keys
of their group, with the causal mask and softmax scale of the model, and
adds it to $s_g$. The score is thus obtained within the single prefill
pass, without a second pass over the prompt, and only one vector of $N$
scores per KV group is kept until the set is built. When a turn adds fewer
than $n_{\text{obs}}$ new tokens, all of them serve as observation queries.

A question rarely attends to the answer it asks for, but to the tokens
that identify it: a key, a speaker name or a function name. The answer
that the model must reproduce---the value of the key, the line of the
speaker, the body of the function---follows these tokens, spans many of
them, and need not resemble the question, so its own score can be low.
LastSet therefore spreads each peak of $s_g$ forward with a max-pool,
\begin{equation}
    s'_g(t) = \max_{t - w \leq u \leq t} s_g(u),
    \label{eq:lastset-pool}
\end{equation}
with window $w$, so that each high-scoring token brings the $w$ tokens
that follow it into the selection. The pool is one-sided, unlike the
symmetric pooling of SnapKV~\cite{li2024snapkv}, which spends half of its
window on the tokens before a peak; a one-sided window of the same width
reaches twice as far into the answer.

\subsubsection{Selected set and decoding}
\label{subsubsec:lastset-set}

For each KV group, LastSet keeps the first $n_{\text{sink}}$ tokens
(attention sinks~\cite{xiao2024efficient}), the $k$ tokens with the
highest $s'_g$ among positions $[n_{\text{sink}}, n - n_{\text{obs}})$, and
a recent window made of the last $n_{\text{obs}}$ prompt tokens and every
token generated in the turn:
\begin{equation}
    \mathcal{I}_g = \{1, \ldots, n_{\text{sink}}\} \,\cup\,
    \operatorname{top}\text{-}k\, s'_g \,\cup\,
    \{n - n_{\text{obs}} + 1, \ldots, N\}.
    \label{eq:lastset-set}
\end{equation}
The sinks and the recent window are kept whatever their score: the sinks
absorb the attention that no other token takes~\cite{xiao2024efficient},
and the window holds the question itself and the answer being written.
The top-$k$ is taken outside them, so the whole budget $k$ goes to entries
chosen by the score. Sets are built per KV group rather than per query
head, because all query heads of a group read the same keys and values
(Definition~\ref{def:ssa}); a group thus reads its selected entries once
for all of its heads.

The same set $\mathcal{I}_g$ serves every layer and every decode step of the
turn, so each step attends to $|\mathcal{I}_g| \ll N$ entries through
Equation~\eqref{eq:sparse-attention}, with no per-step scoring. Sharing the
set across layers loses little: during decoding, the shared set covers
within $3$ points as much of the attention mass of each layer from $4$ to
$27$ as a set chosen by the same rule from that layer's own attention. In
the first four layers attention is spread over the prompt and neither set
covers half of it, yet restricting these layers to the shared set leaves
the loss of the dense answer unchanged.
Sharing the set across decoding steps relies on the answer of a turn
staying on what its question asked for; tasks whose answer must visit
the whole context, such as counting words over a long list, break this
assumption, and we return to them in §\ref{subsec:exp-lastset-acc}.

\subsubsection{Sessions with several turns}
\label{subsubsec:lastset-turns}

LastSet evicts nothing. The set of a turn is only a list of positions; the
keys and values of the whole session, including the answers generated in
earlier turns, stay in the paged KV Cache. When the student asks the next
question, prefix caching reuses this cache and prefills only the new
tokens, whose queries attend to all $N$ cached entries with full
attention, as in dense prefill. Equations~\eqref{eq:lastset-score}--\eqref{eq:lastset-set}
are evaluated on these queries and replace the previous set, so a question
about a part of the material that the previous turn never read is answered
from the same cache, without recomputing it. An eviction method can answer
such a question only after the dropped entries are prefilled again, which
gives up the benefit of prefix caching. Algorithm~\ref{alg:lastset}
summarizes one turn.

\subsubsection{Implementation and cost}
\label{subsubsec:lastset-cost}

LastSet stores, for each request and KV group, the $k$ selected positions
as sorted $32$-bit integers; the sinks and the recent window follow from
the sequence length and are not stored. For Qwen2.5-7B-Instruct-1M, with
$G = 4$ KV groups and $k = 8{,}192$, this is $128$\,KB per request,
against about $7$\,GB for the bf16 KV Cache of a $128$k-token context. At
each decode step, the positions are mapped to slots of the paged cache
through the block table of the request, and each pair of a request and a
KV group is treated as one sequence of a multi-query attention whose
queries are the $|H_g|$ heads of the group. Decoding therefore reuses the
FlashAttention kernel of dense decoding~\cite{dao2022flashattention} and,
since the size of the set is fixed within a turn, runs under CUDA graphs.

At step $j$ of a turn, each layer reads the keys and values of the
$|\mathcal{I}_g|$ entries of Equation~\eqref{eq:lastset-set} for each KV
group instead of all $N$, so attention reads
\begin{equation}
    M_j = 2\, n_L\, G\, d_h\, \beta \left(n_{\text{sink}} + k + n_{\text{obs}} + j\right)
    \label{eq:lastset-cost}
\end{equation}
bytes, with $\beta$ bytes per element, against $2\, n_L\, G\, d_h\, \beta\, N$
for dense decoding. $M_j$ does not depend on $N$: at $N = 256$k and
$k = 8{,}192$, a step reads about $3\%$ of the entries that dense decoding
reads. Every concurrent request reads its own cache, while the weights are
read once for the whole batch, so this saving grows with the number of
requests that share the GPU. Scoring adds one attention pass of
$n_{\text{obs}}$ queries over $N$ keys in half of the layers, once per
turn, which is small next to the prefill of the turn. Per-step selectors
instead pay a scoring pass at every step (Quest) or at their filter layers
(OmniKV), and eviction methods save the same traffic but lose the dropped
entries for later turns.

\begin{algorithm}[!t]
\caption{LastSet: one turn of sparse decoding}
\label{alg:lastset}
\begin{algorithmic}[1]
\Require Prompt of the turn (length $n$); KV Cache of all previous turns; budget $k$; $n_{\text{sink}}$, $n_{\text{obs}}$, $w$
\State Prefill the prompt with full attention; append its keys and values to the KV Cache
\ForAll{KV group $g$}
    \State Compute $s_g$ from the last $n_{\text{obs}}$ queries via Eq.~\eqref{eq:lastset-score}
    \State Pool $s'_g \gets$ Eq.~\eqref{eq:lastset-pool}
    \State $\mathcal{I}_g \gets$ sinks $\cup$ top-$k$ of $s'_g$ $\cup$ recent window \Comment{Eq.~\eqref{eq:lastset-set}}
\EndFor
\While{not end of turn}
    \State Decode one token; every layer attends to $\mathcal{I}_g$ only
    \State Append the new token to the KV Cache and to every $\mathcal{I}_g$
\EndWhile
\State Keep the full KV Cache; the next turn selects again
\end{algorithmic}
\end{algorithm}

\subsection{REPAIR: Self-Improving Playbook Adaptation}
\label{subsec:agentic-engine}

\begin{figure*}[t]
\centering
\includegraphics[width=0.95\textwidth]{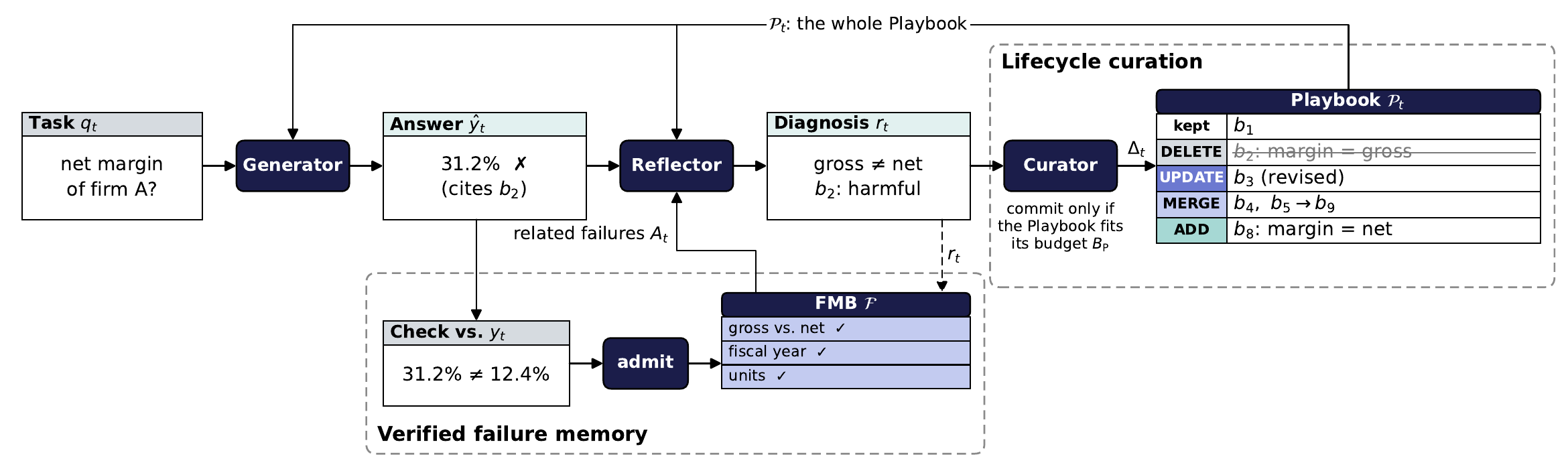}
\caption{REPAIR during offline adaptation, with an illustrative
financial question in which a wrong Playbook rule ($b_2$) leads to a wrong
margin. The Generator, Reflector and Curator form the ACE loop. Lifecycle
curation lets the Curator add, update, merge and delete Playbook entries,
and commits an edit only if the Playbook stays within its budget. The
verified failure memory stores failures confirmed against the reference
and gives the Reflector the earlier failures related to the current task.}
\label{fig:repair}
\end{figure*}

REPAIR builds on the Generator--Reflector--Curator loop of
ACE~\cite{agenticcontext2026} and keeps the model weights fixed. It adds
one part for each observation in
§\ref{subsec:obs-curation}--\ref{subsec:obs-agentic}: the Curator can revise,
merge and delete entries as well as add them
(§\ref{subsubsec:ace-curator}), and the Reflector sees verified earlier
failures related to the current one (§\ref{subsubsec:ace-fmb}). The result
is a Playbook that keeps improving as a reusable system prompt.
Figure~\ref{fig:repair} shows the two parts around the ACE loop.

\subsubsection{Architecture and information flow}
\label{subsubsec:ace-architecture}

With $q_t$, $c_t$ and $y_t$ as in Definition~\ref{def:ace}, the Playbook
is $\mathcal{P}_t=\{b_j\}$ with
$b_j=(\mathrm{id}_j,\mathrm{section}_j,\mathrm{text}_j,c^{+}_j,c^{-}_j)$,
where $c^{+}_j$ and $c^{-}_j$ count the helpful and harmful tags of the
entry. The verified failures $\mathcal{F}_t$, read by the
Reflector, complement it.{} Algorithm~\ref{alg:scale-agentic} gives the offline loop.

\subsubsection{Generation and reflection}
\label{subsubsec:ace-generator}

The Generator receives the task and the whole Playbook, and returns a
trajectory (a calculation, a program or a tool trace), an answer and the
identifiers $U_t$ of the entries it cites:
\begin{equation}
    (\rho_t,\hat y_t,U_t)
    =G_\phi(q_t,c_t,\mathcal{P}_t).
    \label{eq:ace-generator-role}
\end{equation}

\label{subsubsec:ace-reflector}
The Reflector compares the output of Equation~\eqref{eq:ace-generator-role}
with the feedback $y_t$, given the
Playbook and the verified failures $A_t$ retrieved for the task
(§\ref{subsubsec:ace-fmb}):
\begin{equation}
    (r_t,\Gamma_t)
    =R_\phi(q_t,\rho_t,\hat y_t,y_t,U_t,\mathcal{P}_t,A_t),
    \label{eq:ace-reflector-role}
\end{equation}
where $\Gamma_t$ tags each cited entry as helpful, neutral or harmful,
and the diagnosis $r_t=(e_t,z_t,a_t^\star,\kappa_t)$ names the error, its
suspected cause, a corrective approach and a reusable lesson.

\subsubsection{Lifecycle curation}
\label{subsubsec:ace-curator}

Section~\ref{subsec:obs-curation} shows that a Curator that can only add
entries keeps outdated advice next to its corrections. The lifecycle
Curator reads $r_t$, $\Gamma_t$, the Playbook and a set $\mathcal{X}_t$ of
maintenance candidates, and proposes typed edits:
\begin{equation}
    \Delta_t=C_\phi(\mathcal{P}_t,r_t,\Gamma_t,\mathcal{X}_t).
    \label{eq:ace-curator-role}
\end{equation}
\texttt{ADD} inserts a new entry; \texttt{UPDATE} rewrites an entry in
place, keeping its identifier and counters, so that a rule is narrowed or
corrected instead of being contradicted by a new one;
\texttt{CREATE\_META} opens a new section; and \texttt{MERGE} replaces two
or more entries with one written by the Curator. Merge candidates
$\mathcal{C}_t$ are clusters of similar entries over the whole Playbook;
the Curator merges a cluster only if its entries are redundant rather than
complementary (Appendix~\ref{app:repair-impl}). \texttt{DELETE} may target
only entries whose harmful tags clearly outweigh their helpful ones,
\begin{equation}
    \mathcal{X}^{\mathrm{del}}_t
    =\{b_j\in\mathcal{P}_t:
      c^{-}_j\geq c^{-}_{\min},\ c^{-}_j-c^{+}_j\geq\delta_{\mathrm{del}}\},
    \label{eq:ace-delete-candidates}
\end{equation}
and the Curator removes such an entry only with a stated reason, so
$\mathcal{X}_t=\mathcal{C}_t\cup\mathcal{X}^{\mathrm{del}}_t$. The executor
applies the valid edits $\Delta_t^{\mathrm{valid}}\subseteq\Delta_t$ of
Equation~\eqref{eq:ace-curator-role} deterministically through $\Phi$, instantiating
Equation~\eqref{eq:ace-context-update}:
\begin{equation}
    \widetilde{\mathcal{P}}_{t+1}
    =\operatorname{Prune}_t\!
       \left(\Phi(\mathcal{P}_t,\Delta_t^{\mathrm{valid}})\right),
    \label{eq:ace-safe-curation}
\end{equation}
where $\operatorname{Prune}_t$ periodically removes entries that were
never tagged. Under a token budget $B_{\mathrm{P}}$, an oversized
candidate is first compressed by the Curator through \texttt{UPDATE},
\texttt{MERGE} and \texttt{DELETE}, and is committed only if it fits;
otherwise the previous Playbook is kept
(Equation~\eqref{eq:ace-budget-commit}, Appendix~\ref{app:repair-impl}).

\subsubsection{Verified failure memory}
\label{subsubsec:ace-fmb}

Section~\ref{subsec:obs-agentic} shows that the same mistakes recur on
tasks that look different. The Failure Memory Bank (FMB) $\mathcal{F}$
therefore keeps past failures for the Reflector, but only those confirmed
by an external check, since a plausible reflection need not describe a
real error. A record keeps the task, the prediction, the reference, the
diagnosis, a failure type, its evidence and the Curator edits that
followed. It is admitted only if the task checker, never the LLM, finds a
mismatch with the reference and the failure is a task error rather than
an infrastructure fault or an ambiguous input
(Equation~\eqref{eq:ace-fmb-admission}). For a new task, the FMB ranks
records by semantic similarity, overlap with the stored diagnosis and past
usefulness (Equation~\eqref{eq:ace-fmb-retrieval}), and passes the top
$K_{\mathrm{F}}$ to the Reflector as $A_t$ in
Equation~\eqref{eq:ace-reflector-role}.

\subsubsection{Offline adaptation and deployment}
\label{subsubsec:ace-deployment}

Algorithm~\ref{alg:scale-agentic} retrieves memory before writing the
record of the current task, so a task never retrieves its own answer, and
each part can be disabled separately for ablations. At deployment, the
Playbook\ is{} frozen: execution uses
Equation~\eqref{eq:ace-generator-role} with read-only retrieval and no
labels, reflection, curation or memory writes, so adaptation adds no cost
to the response path.

\begin{algorithm}[!t]
\caption{Offline adaptation in REPAIR.}
\label{alg:scale-agentic}
\begin{algorithmic}[1]

\Require Training tasks $\mathcal{D}$; initial $\mathcal{P},\mathcal{F}$; enabled mechanisms; optional budget $B_{\mathrm{P}}$
\Ensure Adapted $\mathcal{P},\mathcal{F}$
\State Check initial Playbook budget if enabled
\For{$(q_t,c_t,y_t)\in\mathcal{D}$}
    \State $(\rho_t,\hat y_t,U_t)\gets G_\phi(q_t,c_t,\mathcal{P})$

    \State $V_t\gets\operatorname{CheckOutcome}(\hat y_t,y_t,\rho_t)$
    \State $A_t\gets\operatorname{RetrieveVerified}(\mathcal{F},q_t)$
    \State $(r_t,\Gamma_t)\gets R_\phi(q_t,\rho_t,\hat y_t,y_t,U_t,\mathcal{P},A_t)$
    \State Update usage counters using $\Gamma_t$ under the active budget guard
    \State $f_t\gets\operatorname{BuildFailure}(r_t,V_t,q_t,\hat y_t,y_t)$
    \If{$f_t$ passes Eq.~\eqref{eq:ace-fmb-admission}}
        \State $\mathcal{F}\gets\mathcal{F}\cup\{f_t\}$
    \EndIf
    \State $\mathcal{X}_t\gets\operatorname{MaintenanceCandidates}(\mathcal{P})$
    \State $\Delta_t\gets C_\phi(\mathcal{P},r_t,\Gamma_t,\mathcal{X}_t)$
    \State Construct candidate via Eq.~\eqref{eq:ace-safe-curation}
    \If{budget enforcement is enabled}
        \State Maintain candidate for at most $R_{\mathrm{bud}}$ attempts
        \State Commit or roll back via Eq.~\eqref{eq:ace-budget-commit}
    \Else
        \State Accept the candidate Playbook
    \EndIf
    \State Attach Curator outcomes to the admitted failure, if any

\EndFor

\State \Return $\mathcal{P},\mathcal{F}$

\end{algorithmic}
\end{algorithm}

\section{Experiments and Results}
\label{sec:experiments}

We evaluate DeepEdu-F in two parts. Because the two components target
different bottlenecks and call for different baselines, we study each on
its own: the long-context inference engine (§\ref{subsec:exp-lastset}) is
measured against sparse decoding methods on long-context benchmarks,
whereas the self-improving agentic layer (§\ref{subsec:exp-agentic}) is
evaluated as a controlled ablation over its mechanisms on agentic
reasoning benchmarks.

\subsection{Long-Context Inference Engine}
\label{subsec:exp-lastset}

We evaluate LastSet (§\ref{subsec:long-context-engine}) against four sparse
decoding methods at the same token budget. We aim to answer three research
questions:
\begin{itemize}
    \item \textbf{RQ1}: Does LastSet reduce the time per output token
    compared to dense and sparse decoding, across context lengths and
    numbers of concurrent requests?
    \item \textbf{RQ2}: Does selecting once per turn preserve accuracy on
    long-context benchmarks at the same budget as the baselines?
    \item \textbf{RQ3}: How does accuracy depend on the budget $k$?
\end{itemize}

\subsubsection{Experimental Setup}
\label{subsec:exp-setup}

\paragraph{Hardware, software and model}
All experiments run on a single NVIDIA H100 80\,GB GPU. All methods are
implemented in the same serving engine, SparseEngine~\cite{hao2026sparseengine},
with bf16 weights, prefix caching and CUDA graphs, so that
they differ only in the decode attention. The model is
Qwen2.5-7B-Instruct-1M~\cite{yang2025qwen251m}. LastSet uses
$n_{\text{sink}} = n_{\text{obs}} = 128$ and a pooling window of $w = 64$,
and decodes with the kernel of §\ref{subsubsec:lastset-cost}.

\paragraph{Baselines}
We compare with two KV eviction methods, SnapKV~\cite{li2024snapkv}
(pooling kernel $7$) and AdaKV~\cite{feng2024adakv}, which keep the
tokens selected at prefill and drop the rest; Quest~\cite{tang2024quest},
which keeps the full cache and selects pages at every decode step; and
OmniKV~\cite{omnikv2025}, which selects tokens at a few filter layers and
shares them with the other layers. All use the same budget of $k$
selected tokens, and dense decoding is the reference.

\paragraph{Benchmarks and metrics}
On InfiniteBench~\cite{zhang_etal_2024_bench} we use four tasks:
kv\_retrieval, longdialogue\_qa\_eng, math\_find and code\_debug. On
RULER~\cite{ruler2024} we use its thirteen tasks at a context length of
$128$k tokens. Both use their official prompts, generation lengths and
metrics, on samples drawn with a fixed seed. All methods
use a budget of $k = 8{,}192$; for LastSet we also vary
$k \in \{2048, 4096, 8192\}$ on InfiniteBench.
Time per output token (TPOT) is measured on $256$ output tokens at
contexts of $32$k, $128$k and $256$k tokens with $1$ to $16$ concurrent
requests.

\subsubsection{Decoding Speed}
\label{subsec:exp-tpot}

\begin{figure*}[t]
\centering
\includegraphics[width=0.95\textwidth]{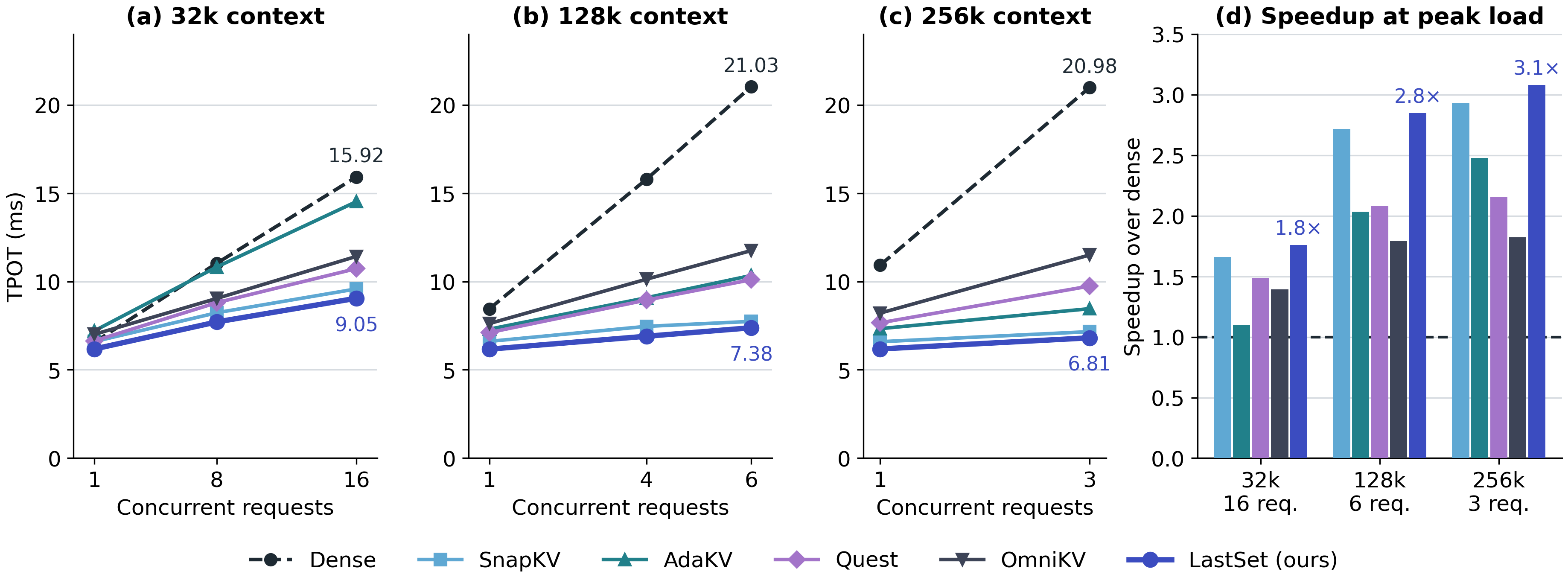}
\caption{(a)--(c) Time per output token versus the number of concurrent
requests at three context lengths. (d) Speedup over dense decoding of every
sparse method at the highest load of each context length.}
\label{fig:lastset-tpot}
\end{figure*}

Table~\ref{tab:lastset-tpot} and Figure~\ref{fig:lastset-tpot} answer RQ1.
LastSet is the fastest method in every setting, and its advantage over
dense decoding grows with the context and the load, from $1.8\times$ at
$32$k tokens to $3.1\times$ at $256$k (Figure~\ref{fig:lastset-tpot}d).
Equation~\eqref{eq:lastset-cost} explains this trend: with one request, the
time per token of LastSet stays the same from $32$k to $256$k tokens,
because the entries it reads do not grow with the context, whereas dense
decoding slows down as the cache grows. LastSet is also faster than Quest
and OmniKV, which score the cache during decoding, and than SnapKV, which
reads a set of the same size but drops every entry outside it
(§\ref{subsec:obs-tradeoff}).

\begin{table*}[t]
\centering
\caption{Time per output token (ms) with Qwen2.5-7B-Instruct-1M on one H100,
by context length and number of concurrent requests ($256$ output tokens,
budget $8{,}192$). Best non-dense value per column in bold.}
\label{tab:lastset-tpot}
\small
\setlength{\tabcolsep}{4pt}
\begin{tabular*}{\textwidth}{@{\extracolsep{\fill}}lcccccccc@{}}
\toprule
 & \multicolumn{3}{c}{\textbf{32k}} & \multicolumn{3}{c}{\textbf{128k}} & \multicolumn{2}{c}{\textbf{256k}} \\
\cmidrule(lr){2-4}\cmidrule(lr){5-7}\cmidrule(lr){8-9}
\textbf{Method} & 1 & 8 & 16 & 1 & 4 & 6 & 1 & 3 \\
\midrule
Dense                         & 6.56 & 11.03 & 15.92 & 8.45 & 15.79 & 21.03 & 10.92 & 20.98 \\
\midrule
SnapKV~\cite{li2024snapkv}    & 6.60 & 8.23 & 9.58 & 6.61 & 7.46 & 7.74 & 6.59 & 7.16 \\
AdaKV~\cite{feng2024adakv}    & 7.21 & 10.82 & 14.52 & 7.30 & 9.08 & 10.34 & 7.33 & 8.46 \\
Quest~\cite{tang2024quest}    & 6.64 & 8.80 & 10.73 & 7.11 & 8.95 & 10.10 & 7.66 & 9.74 \\
OmniKV~\cite{omnikv2025}      & 7.01 & 9.04 & 11.42 & 7.62 & 10.13 & 11.75 & 8.21 & 11.50 \\
\textbf{LastSet (ours)}       & \textbf{6.18} & \textbf{7.71} & \textbf{9.05} & \textbf{6.17} & \textbf{6.90} & \textbf{7.38} & \textbf{6.18} & \textbf{6.81} \\
\bottomrule
\end{tabular*}
\end{table*}

\subsubsection{Accuracy on InfiniteBench and RULER}
\label{subsec:exp-lastset-acc}

Tables~\ref{tab:lastset-ib} and~\ref{tab:lastset-ruler} answer RQ2. On both
benchmarks, LastSet has the highest average among the sparse methods and
stays within two points of dense decoding. Its advantage lies in
retrieval. It matches dense decoding on Retr.KV, where every other sparse
method loses at least $20$ points, and on all eight retrieval tasks of
RULER, including the long UUID needles of S3 and MK3 on which the eviction
methods fail. In these tasks the question names a key and the answer is
the long value that follows it, which the forward pool of
Equation~\eqref{eq:lastset-pool} keeps together with the key. Its weak point
is aggregation: on CWE the answer counts words over the whole context, so
no single set chosen from the question covers enough of their occurrences,
and OmniKV, which selects again at every step, does better
(§\ref{subsubsec:lastset-set}).

The lower rows of Table~\ref{tab:lastset-ib} answer RQ3. Halving the budget
to $k = 4{,}096$ leaves the average of LastSet unchanged, and at
$k = 2{,}048$ LastSet still outperforms every baseline at $k = 8{,}192$.
Since the traffic of a decode step is proportional to the size of the set
(Equation~\eqref{eq:lastset-cost}), a smaller budget gives a faster decoder
at little cost in accuracy.

\begin{table}[t]
\centering
\caption{InfiniteBench with Qwen2.5-7B-Instruct-1M. Upper rows: all methods
at budget $k = 8{,}192$, best non-dense value per column in bold. Lower
rows: LastSet with smaller budgets. Avg: mean over the four tasks.}
\label{tab:lastset-ib}
\footnotesize
\begin{tabular*}{\columnwidth}{@{\extracolsep{\fill}}lccccc@{}}
\toprule
\textbf{Method} & \textbf{Retr.KV} & \textbf{En.Dia} & \textbf{Math} & \textbf{Code} & \textbf{Avg} \\
\midrule
Dense & 56.7 & 26.7 & 40.0 & 23.3 & 36.7 \\
\midrule
SnapKV~\cite{li2024snapkv} & 6.7 & 13.3 & \textbf{40.0} & 26.7 & 21.7 \\
AdaKV~\cite{feng2024adakv} & 10.0 & \textbf{20.0} & \textbf{40.0} & 26.7 & 24.2 \\
Quest~\cite{tang2024quest} & 10.0 & 13.3 & 36.7 & \textbf{30.0} & 22.5 \\
OmniKV~\cite{omnikv2025} & 36.7 & \textbf{20.0} & \textbf{40.0} & 26.7 & 30.8 \\
\textbf{LastSet (ours)} & \textbf{56.7} & 16.7 & \textbf{40.0} & 26.7 & \textbf{35.0} \\
\midrule
\quad $k = 4{,}096$ & 56.7 & 16.7 & 40.0 & 26.7 & 35.0 \\
\quad $k = 2{,}048$ & 60.0 & 13.3 & 36.7 & 26.7 & 34.2 \\
\bottomrule
\end{tabular*}
\end{table}

\begin{table*}[t]
\centering
\caption{RULER (thirteen tasks, $128$k tokens) with Qwen2.5-7B-Instruct-1M at
budget $k = 8{,}192$. S: single needle, MK: multi-key, MV: multi-value, MQ:
multi-query NIAH; VT: variable tracking; CWE / FWE: common / frequent word
extraction. Avg: mean over the thirteen tasks. Best non-dense value per
column in bold.}
\label{tab:lastset-ruler}
\footnotesize
\setlength{\tabcolsep}{3pt}
\begin{tabular*}{\textwidth}{@{\extracolsep{\fill}}lcccccccccccccc@{}}
\toprule
 & \multicolumn{8}{c}{\textbf{Retrieval (NIAH)}} & \textbf{Tracing} & \multicolumn{2}{c}{\textbf{Aggregation}} & \multicolumn{2}{c}{\textbf{QA}} & \\
\cmidrule(lr){2-9}\cmidrule(lr){10-10}\cmidrule(lr){11-12}\cmidrule(lr){13-14}
\textbf{Method} & S1 & S2 & S3 & MK1 & MK2 & MK3 & MV & MQ & VT & CWE & FWE & QA1 & QA2 & \textbf{Avg} \\
\midrule
Dense & 100.0 & 100.0 & 100.0 & 100.0 & 93.3 & 86.7 & 84.2 & 99.2 & 89.3 & 30.0 & 82.2 & 50.0 & 30.0 & 80.4 \\
\midrule
SnapKV~\cite{li2024snapkv} & \textbf{100.0} & \textbf{100.0} & 46.7 & \textbf{100.0} & \textbf{93.3} & 6.7 & 86.7 & 99.2 & 91.3 & 18.7 & 60.0 & \textbf{53.3} & \textbf{36.7} & 68.7 \\
AdaKV~\cite{feng2024adakv} & \textbf{100.0} & \textbf{100.0} & 36.7 & 96.7 & 90.0 & 6.7 & 84.2 & 99.2 & 90.7 & 20.0 & 65.6 & \textbf{53.3} & \textbf{36.7} & 67.7 \\
Quest~\cite{tang2024quest} & \textbf{100.0} & \textbf{100.0} & \textbf{100.0} & \textbf{100.0} & \textbf{93.3} & 60.0 & 82.5 & 98.3 & 86.0 & 16.0 & \textbf{73.3} & 50.0 & 30.0 & 76.1 \\
OmniKV~\cite{omnikv2025} & \textbf{100.0} & \textbf{100.0} & \textbf{100.0} & \textbf{100.0} & 86.7 & 73.3 & 82.5 & \textbf{100.0} & 90.0 & \textbf{40.7} & 70.0 & 50.0 & 33.3 & 79.0 \\
\textbf{LastSet (ours)} & \textbf{100.0} & \textbf{100.0} & \textbf{100.0} & \textbf{100.0} & \textbf{93.3} & \textbf{86.7} & \textbf{94.2} & \textbf{100.0} & \textbf{92.0} & 6.3 & 71.1 & 50.0 & \textbf{36.7} & \textbf{79.3} \\
\bottomrule
\end{tabular*}
\end{table*}

\subsection{Self-Improving Agentic Layer}
\label{subsec:exp-agentic}

We evaluate REPAIR (\S\ref{subsec:agentic-engine}) independently of
LastSet, using the same backbone for ACE (ADD), lifecycle curation,
verified failure memory, and their combination. We ask the following
research questions:
\begin{itemize}
    \item \textbf{RQ4}: Do individual lifecycle operations and their
    full combination improve on the ADD-oriented ACE baseline?
    \item \textbf{RQ5}: Does grounding the Reflector in verified prior
    failures improve ACE without lifecycle extensions?
    \item \textbf{RQ6}: Does combining lifecycle curation with verified
    FMB improve on either component alone across financial reasoning
    and interactive-agent tasks?

\end{itemize}

\subsubsection{Experimental Setup}
\label{subsec:agentic-experiments}

\paragraph{Hardware, software and model}
Generator, Reflector, and Curator use Qwen3-4B-Instruct-2507, served
through vLLM on a single NVIDIA H100 with $80$\,GB HBM3. On Formula
and FiNER, all variants use one epoch, at most three agentic rounds
per sample, a $2{,}048$-token output limit, and seed $42$. We evaluate
every $50$ training steps, save checkpoints every $25$ steps, and use
five test workers. The Curator runs after each training sample.
AppWorld retains the same backbone, prompts, task order, seeds, and
split-specific rollout limits across its ablations.

\paragraph{Benchmarks and supervision}
All adaptation is offline, with Ground Truth feedback available during
training. Formula uses $500$ training, $300$ validation, and $200$ test
examples; the lifecycle and FMB studies use full FiNER
(\texttt{finer}), with $1{,}000$ training, $500$ validation, and $441$
test examples. Finance checkpoints are selected by validation accuracy.
On AppWorld, training uses reference execution feedback, and each
trained Playbook is evaluated separately on \texttt{test\_normal}
and \texttt{test\_challenge}. The Playbook is frozen before testing;
test answers and trajectories never update it or the failure memory.

\paragraph{Baseline and ablations}
The baseline is the ACE Generator--Reflector--Curator
architecture~\cite{agenticcontext2026}, configured with an ADD-only
Curator and no failure memory. This control isolates the extensions;
it does not imply that ACE's grow-and-refine design lacks cleanup.
Every variant starts from the same initial Playbook within a benchmark.
Each table row independently extends ACE (ADD), rather than adding
to the preceding row. UPDATE and MERGE enable one additional operation;
DELETE+PRUNE combines harmful-entry removal with periodic unused-entry
pruning. Full ops enables ADD, UPDATE, MERGE, DELETE, and PRUNE.
The FMB row adds verified failure memory without lifecycle extensions,
while the final row enables both. Automatic post-Curator
BulletpointAnalyzer cleanup is disabled throughout to avoid confounding
the operation comparison.

\paragraph{Lifecycle and memory settings}
MERGE uses DBSCAN candidates with similarity threshold
$\eta_{\mathrm{merge}}=0.90$, cosine-distance radius
$\varepsilon=0.10$, and minimum neighborhood size $m=2$.
DELETE uses $c^{-}_{\min}=3$ and $\delta_{\mathrm{del}}=4$;
PRUNE runs every $T_{\mathrm{prune}}=50$ training samples.
The prompt specifies a $10{,}000$-token Playbook budget, but hard
enforcement is disabled: overflow repair could otherwise invoke
UPDATE and MERGE in a run intended to isolate another operation.

Both FMB presets retrieve up to $K_{\mathrm{F}}=5$ records for the
Reflector. Admission uses confidence threshold
$\eta_{\mathrm{ver}}=0.8$; retrieval forms a pool of up to
$4K_{\mathrm{F}}$ semantic candidates, applies relevance cutoff $0.2$,
and uses weights $(0.65,0.25,0.10)$ for semantic similarity,
diagnostic token overlap, and recorded usefulness.
Failure verification uses dataset references on finance and
execution-based evaluation on AppWorld. Record construction and
retrieval follow \S\ref{subsubsec:ace-fmb} and
Appendix~\ref{app:repair-impl}; no legacy-memory variant is included.

\paragraph{Metrics}
We report final test accuracy, computed per label on FiNER and per
answer on Formula, and their unweighted mean. AppWorld reports Task
Goal Completion (TGC) and Scenario Goal Completion (SGC) on both
test splits; its average is the unweighted mean of these four
percentages. These averages summarize performance but do not replace
the individual metrics. Results are descriptive comparisons without
multi-seed uncertainty estimates; the tables omit initial-test and
best-validation scores.

\subsubsection{Financial Reasoning: Formula and FiNER}

Table~\ref{tab:agentic-finance-ablation} answers RQ4 and RQ5. Every
lifecycle operation improves on ACE (ADD) on both benchmarks, MERGE the
most, but enabling all of them together (Full ops) gains less than MERGE
alone: more editing actions do not by themselves produce better guidance.
Verified FMB alone also improves both benchmarks. Combining it with full
lifecycle curation gives the best average, $70.10\%$ against $61.80\%$
for ACE (ADD), and recovers what full ops alone loses (RQ6). Reflection
grounded in verified failures thus helps most when the Curator has several
ways to revise the Playbook. The combination is not best everywhere,
however: MERGE alone remains slightly better on FiNER.

\begin{table}[t]
\centering
\caption{Final test accuracy (\%) on FiNER and Formula after offline
adaptation with Ground Truth. Operation extensions retain ADD.
Full ops includes PRUNE. All variants use the same
backbone and training setup. FMB denotes verified memory with $K_{\mathrm{F}}=5$.
Bold marks the highest value in each column.}
\label{tab:agentic-finance-ablation}
\footnotesize
\setlength{\tabcolsep}{3pt}
\begin{tabular*}{\columnwidth}{@{\extracolsep{\fill}}lccc@{}}
\toprule
Method & FiNER$\uparrow$ & Formula$\uparrow$ & Avg.$\uparrow$ \\
\midrule
ACE (ADD)~\cite{agenticcontext2026} & 53.60 & 70.00 & 61.80 \\
\midrule
\multicolumn{4}{l}{\textit{Lifecycle operations}} \\
\quad + UPDATE & 56.00 & 76.00 & 66.00 \\
\quad + MERGE & \textbf{60.00} & 78.00 & 69.00 \\
\quad + DELETE + PRUNE & 59.47 & 73.00 & 66.24 \\
\quad + Full ops & 54.00 & 74.00 & 64.00 \\
\midrule
\multicolumn{4}{l}{\textit{Failure memory and joint configurations}} \\
\quad + FMB & 57.90 & 78.00 & 67.95 \\
\quad + Full ops + FMB & 59.70 & \textbf{80.50} & \textbf{70.10} \\
\bottomrule
\end{tabular*}
\end{table}

\subsubsection{Interactive Agents: AppWorld}

Table~\ref{tab:agentic-appworld-ablation} shows the same pattern on
stateful, tool-using tasks. Among single operations, DELETE+PRUNE helps
most, and enabling every operation is again worse than the best single one
(RQ4). FMB alone raises the average but leaves challenge SGC unchanged
(RQ5). The combination improves all four metrics over both FMB alone and
full ops, and raises the average from $9.28$ for ACE (ADD) to $16.28$
(RQ6). Together with the finance results, this supports combining
failure-grounded reflection with lifecycle maintenance rather than relying
on more operations alone. The challenge split remains hard, however: its
SGC stays below $3\%$ for every variant, so these gains do not resolve the
harder interactive tasks.

\begin{table*}[t]
\centering
\caption{Aggregate AppWorld performance (\%) after offline adaptation
with Ground Truth execution feedback. Each frozen Playbook is tested
on both splits. Avg.\ is the mean of the four reported metrics.
FMB denotes verified memory with $K_{\mathrm{F}}=5$. Bold marks the
highest value in each column.}
\label{tab:agentic-appworld-ablation}
\small
\setlength{\tabcolsep}{4pt}
\begin{tabular*}{0.9\textwidth}{@{\extracolsep{\fill}}lccccc@{}}
\toprule
Method & \multicolumn{2}{c}{Test-Normal}
       & \multicolumn{2}{c}{Test-Challenge} & Avg.$\uparrow$ \\
\cmidrule(lr){2-3}\cmidrule(lr){4-5}
 & TGC$\uparrow$ & SGC$\uparrow$ & TGC$\uparrow$ & SGC$\uparrow$ & \\
\midrule
ACE (ADD)~\cite{agenticcontext2026} & 20.8 & 8.9 & 6.7 & 0.7 & 9.28 \\
\midrule
\multicolumn{6}{l}{\textit{Lifecycle operations}} \\
\quad + UPDATE & 21.4 & 10.7 & 8.9 & 2.2 & 10.80 \\
\quad + MERGE & 23.2 & 8.9 & 10.1 & \textbf{2.9} & 11.28 \\
\quad + DELETE + PRUNE & 28.6 & 17.9 & 9.3 & 1.4 & 14.30 \\
\quad + Full ops & 24.4 & 14.3 & 7.0 & 1.4 & 11.78 \\
\midrule
\multicolumn{6}{l}{\textit{Failure memory and joint configurations}} \\
\quad + FMB & 26.8 & 10.7 & 7.2 & 0.7 & 11.35 \\
\quad + Full ops + FMB & \textbf{31.6} & \textbf{19.6} & \textbf{11.0} & \textbf{2.9} & \textbf{16.28} \\
\bottomrule
\end{tabular*}
\end{table*}

\section{Conclusion}
\label{sec:conclusion}
We presented DeepEdu-F, a framework for DeepEdu, an AI-tutoring system for
Vietnamese education deployable on the constrained, on-premise hardware
that data-sovereignty regulations require. DeepEdu-F couples two
components targeting the two bottlenecks such deployments face.
\emph{LastSet} attacks the decoding cost of long contexts by selecting the
critical KV Cache entries once per turn, decoding the whole turn from that
set, and keeping the full cache so that every later question can select
again. \emph{REPAIR}, the self-improving agentic layer, attacks the semantic

bottleneck: rather than costly weight updates, it continuously curates an
evolving Playbook through full lifecycle operations and grounds reflection
in a verified Failure Memory Bank. Lifecycle curation revises, consolidates,
and removes instructions, with periodic pruning and optional hard-budget
maintenance; verified FMB supplies reference- or execution-verified
failures to the Reflector.

Our evaluation shows that both components are effective. LastSet is the
fastest decoder in every setting we measured, up to $3.1\times$ faster than
dense decoding at a $256$k-token context, while staying within $1.7$ points
of dense accuracy on InfiniteBench and $1.1$ on RULER and above all four
sparse baselines on average; tasks that aggregate over the whole context
remain its main weakness.

REPAIR provides structured, auditable Playbook maintenance
without fine-tuning, with reported agentic accuracy increasing from
$70.0\%$ to $80.5\%$ on complex tasks.

One main limitation frames our future work: our experiments validate the
enabling mechanisms on established long-context and agentic benchmarks,
whereas evaluating the localized grounding claim directly on native,
region-specific curricula---the setting that ultimately motivates
DeepEdu---remains important future work
toward democratizing trustworthy AI tutoring in resource-constrained
regions.

\appendices
\section{Implementation Details of REPAIR}
\label{app:repair-impl}

\paragraph{Lifecycle edits}
Each edit type can be enabled separately. \texttt{MERGE} candidates
$\mathcal{C}_t$ are the clusters found by DBSCAN~\cite{ester1996density}
over the normalized embeddings $\mathrm{Enc}(\mathrm{text}_j)$ of all
entries, not only those cited for the current task, under cosine distance
with radius $\varepsilon=1-\eta_{\mathrm{merge}}$ for a similarity
threshold $\eta_{\mathrm{merge}}$ and minimum neighborhood size $m$. A
cluster is only a suggestion: the Curator decides whether to merge, writes
the replacement and names its sources, whose counters are summed into the
new entry. For \texttt{DELETE}, the Curator inspects each candidate of
Equation~\eqref{eq:ace-delete-candidates} together with the reflection, and
requests outside that set are rejected. Every edit is checked for a valid
type, the required fields and an existing target; rejected edits are
logged without being applied. $\operatorname{Prune}_t$ in
Equation~\eqref{eq:ace-safe-curation} is the identity except every
$T_{\mathrm{prune}}$ training samples, when it removes the entries with
$c^{+}_j=c^{-}_j=0$; unlike \texttt{DELETE}, it needs no harmful evidence,
so it is evaluated as a separate option.

\paragraph{Playbook budget}
Under a hard budget, the candidate $\widetilde{\mathcal{P}}_{t+1}$ of
Equation~\eqref{eq:ace-safe-curation} must fit $B_{\mathrm{P}}$ tokens,
counted with a fixed tokenizer over the serialized Playbook, including
headers, identifiers and counters. A candidate within budget is accepted.
Otherwise the Curator compresses it through \texttt{UPDATE},
\texttt{MERGE} and, when enabled, \texttt{DELETE} within
Equation~\eqref{eq:ace-delete-candidates}, so that new knowledge is folded
into existing entries instead of being dropped; an edit counts only if it
reduces the token count. After at most $R_{\mathrm{bud}}$ attempts, the
result $\widehat{\mathcal{P}}_{t+1}$ (equal to
$\widetilde{\mathcal{P}}_{t+1}$ when no compression is needed) is committed
only if it fits:
\begin{equation}
    \mathcal{P}_{t+1}=
    \begin{cases}
      \widehat{\mathcal{P}}_{t+1},
        & \operatorname{Tok}
          (\widehat{\mathcal{P}}_{t+1})\leq B_{\mathrm{P}},\\
      \mathcal{P}_t, & \text{otherwise}.
    \end{cases}
    \label{eq:ace-budget-commit}
\end{equation}
A rollback restores the previous Playbook together with its identifier
state, and counter updates pass through the same check, so every committed
Playbook fits the budget.

\paragraph{Failure records}
Each record $f_i$ of the Failure Memory Bank keeps an identifier, the
task $q_i$, the prediction $\hat y_i$, the reference $y_i$, the diagnosis
$r_i$, the failure type $\ell_i$, the evidence $\mathcal{B}_i$, the
verifier $V_i$ with its verification flag and confidence, the cited
Playbook entries $U_i$, the subsequent Curator edits $\Delta_i$ with their
status, and its retrieval and reuse counters.

\paragraph{Admission gate}
A record $f_i$ is admitted when
\begin{equation}
\begin{split}
    g_i={}&\mathbf{1}[\mathrm{verified}(V_i)]\,
            \mathbf{1}[\mathrm{conf}(V_i)\geq\eta_{\mathrm{ver}}]\\
         &{}\cdot\mathbf{1}[\ell_i\in\mathcal{L}]\,
            \mathbf{1}[|\mathcal{B}_i|>0]\\
         &{}\cdot\mathbf{1}[\mathrm{reference\ present}]\,
            \mathbf{1}[\mathrm{observed\ mismatch}]=1,
\end{split}
    \label{eq:ace-fmb-admission}
\end{equation}
where $\mathcal{L}$ contains the task errors: Playbook gaps and
misapplications, and reasoning, calculation, retrieval, verification and
instruction-following errors. Infrastructure failures, malformed outputs
without a verified error and ambiguous inputs are excluded. The check never
asks the LLM for the reference: on the finance tasks it compares the answer
with the dataset reference through the task checker, and on AppWorld it
uses execution outcomes and task-test reports.

\paragraph{Retrieval score}
Retrieval considers the records admitted by
Equation~\eqref{eq:ace-fmb-admission} that are in an active verified,
curated or validated state,
takes up to $4K_{\mathrm{F}}$ semantic candidates and reranks them by
\begin{equation}
    s_i(q)=w_{\mathrm{sem}}\,s_{\mathrm{sem}}(q,f_i)
          +w_{\mathrm{lex}}\,s_{\mathrm{lex}}(q,f_i)
          +w_{\mathrm{use}}\,s_{\mathrm{use}}(f_i),
    \label{eq:ace-fmb-retrieval}
\end{equation}
where $s_{\mathrm{sem}}$ is the normalized embedding similarity,
$s_{\mathrm{lex}}$ the token-set overlap with the stored diagnosis, and
$s_{\mathrm{use}}=(u^{+}_i+1)/(u^{+}_i+u^{-}_i+2)$ the smoothed share of
helpful reuses among the $u^{+}_i$ helpful and $u^{-}_i$ harmful ones. The top $K_{\mathrm{F}}$ records above a relevance cutoff form
$A_t$ in Equation~\eqref{eq:ace-reflector-role}; retrieved failures inform the Reflector, not the Generator.

\IEEEtriggeratref{68}
\bibliography{references}

\begin{thebibliography}{10}
\providecommand{\url}[1]{#1}
\csname url@samestyle\endcsname
\providecommand{\newblock}{\relax}
\providecommand{\bibinfo}[2]{#2}
\providecommand{\BIBentrySTDinterwordspacing}{\spaceskip=0pt\relax}
\providecommand{\BIBentryALTinterwordstretchfactor}{4}
\providecommand{\BIBentryALTinterwordspacing}{\spaceskip=\fontdimen2\font plus
\BIBentryALTinterwordstretchfactor\fontdimen3\font minus
  \fontdimen4\font\relax}
\providecommand{\BIBforeignlanguage}[2]{{%
\expandafter\ifx\csname l@#1\endcsname\relax
\typeout{** WARNING: IEEEtran.bst: No hyphenation pattern has been}%
\typeout{** loaded for the language `#1'. Using the pattern for}%
\typeout{** the default language instead.}%
\else
\language=\csname l@#1\endcsname
\fi
#2}}
\providecommand{\BIBdecl}{\relax}
\BIBdecl

\bibitem{touvron2023llama}
H.~Touvron, L.~Martin, K.~Stone \emph{et~al.}, ``Llama 2: Open foundation and
  fine-tuned chat models,'' \emph{arXiv preprint}, 2023.

\bibitem{bai2023qwen}
J.~Bai, S.~Bai, Y.~Chu \emph{et~al.}, ``Qwen technical report,'' \emph{arXiv
  preprint}, 2023.

\bibitem{yao2023react}
S.~Yao, J.~Zhao, D.~Yu \emph{et~al.}, ``React: Synergizing reasoning and acting
  in language models,'' in \emph{International Conference on Learning
  Representations (ICLR)}, 2023.

\bibitem{llmeducation2026}
S.~Wang, T.~Xu, H.~Li \emph{et~al.}, ``Large language models for education: A
  survey and outlook,'' \emph{IEEE Signal Processing Magazine}, vol.~42, no.~6,
  pp. 51--63, 2025.

\bibitem{vietnam_decree53}
{Government of Vietnam}, ``Decree no. 53/2022/nd-cp detailing a number of
  articles of the law on cybersecurity,''
  \url{https://luatvietnam.vn/an-ninh-quoc-gia/nghi-dinh-53-2022-nd-cp-228170-d1.html},
  2022, introduces cybersecurity and data governance requirements in Vietnam,
  including conditions for data localization and local data storage obligations
  for certain domestic and foreign digital service providers.

\bibitem{cahyawijaya2024llms}
S.~Cahyawijaya, H.~Lovenia, and P.~Fung, ``{LLM}s are few-shot in-context
  low-resource language learners,'' in \emph{Proceedings of the 2024 Conference
  of the North American Chapter of the Association for Computational
  Linguistics: Human Language Technologies (Volume 1: Long Papers)}, 2024.

\bibitem{vnexpress_chatgpt_education_2023}
VnExpress, ``Teachers and students enjoy experimenting with chatgpt,''
  \url{https://vnexpress.net/giao-vien-hoc-sinh-thich-thu-trai-nghiem-chatgpt-4565643.html},
  2023, highlights the increasing adoption of AI assistants by Vietnamese
  students and teachers, alongside concerns over factual inconsistencies.

\bibitem{dnto_chatgpt_2023}
H.~Trang, ``The dangers of using chatgpt to distort vietnamese history,''
  \url{https://www.doanhnhantrevietnam.vn/nguy-hai-khi-ung-dung-chatgpt-xuyen-tac-ve-lich-su-viet-nam-d18556.html},
  2023, details severe historical hallucinations by ChatGPT, such as conflating
  Quang Trung and Nguyen Hue as different kings or fabricating facts about
  national leaders.

\bibitem{lin2024awq}
J.~Lin, J.~Tang, H.~Tang \emph{et~al.}, ``Awq: Activation-aware weight
  quantization for on-device llm compression and acceleration,'' in
  \emph{Proceedings of Machine Learning and Systems}, 2024, pp. 87--100.

\bibitem{frantar2023gptq}
E.~Frantar, S.~Ashkboos, T.~Hoefler, and D.~Alistarh, ``{GPTQ}: Accurate
  post-training quantization for generative pre-trained transformers,'' in
  \emph{The Eleventh International Conference on Learning Representations},
  2023.

\bibitem{kwon2023efficient}
W.~Kwon, Z.~Li, S.~Zhuang \emph{et~al.}, ``Efficient memory management for
  large language model serving with pagedattention,'' in \emph{Symposium on
  Operating Systems Principles (SOSP)}, 2023.

\bibitem{zheng2024sglang}
L.~Zheng, L.~Yin, Z.~Xie \emph{et~al.}, ``{SGLang}: Efficient execution of
  structured language model programs,'' in \emph{Advances in Neural Information
  Processing Systems (NeurIPS)}, 2024.

\bibitem{characterai2024inference}
{Character.AI}, ``Optimizing {AI} inference at {Character.AI},''
  \url{https://blog.character.ai/optimizing-ai-inference-at-character-ai-2/},
  2024.

\bibitem{wang2025kvcachewild}
J.~Wang, J.~Han, X.~Wei \emph{et~al.}, ``{KVCache} cache in the wild:
  Characterizing and optimizing {KVCache} cache at a large cloud provider,'' in
  \emph{Proceedings of the 2025 USENIX Annual Technical Conference (USENIX
  ATC)}, 2025, pp. 465--482.

\bibitem{ji2025manus}
Y.~Ji, ``Context engineering for {AI} agents: Lessons from building {Manus},''
  \url{https://manus.im/blog/Context-Engineering-for-AI-Agents-Lessons-from-Building-Manus},
  2025.

\bibitem{dao2022flashattention}
T.~Dao, D.~Fu, S.~Ermon, A.~Rudra, and C.~R\'{e}, ``Flashattention: Fast and
  memory-efficient exact attention with io-awareness,'' in \emph{Advances in
  Neural Information Processing Systems (NeurIPS)}, 2022, pp. 16\,344--16\,359.

\bibitem{tang2024quest}
J.~Tang, Y.~Zhao, K.~Zhu \emph{et~al.}, ``Quest: query-aware sparsity for
  efficient long-context llm inference,'' in \emph{Proceedings of the 41st
  International Conference on Machine Learning}, 2024.

\bibitem{omnikv2025}
J.~Hao, Y.~Zhu, T.~Wang \emph{et~al.}, ``Omnikv: Dynamic context selection for
  efficient long-context llms,'' in \emph{International Conference on Learning
  Representations}, 2025, pp. 87\,443--87\,464.

\bibitem{li2024snapkv}
Y.~Li, Y.~Huang, B.~Yang \emph{et~al.}, ``Snapkv: Llm knows what you are
  looking for before generation,'' in \emph{Advances in Neural Information
  Processing Systems}, 2024, pp. 22\,947--22\,970.

\bibitem{feng2024adakv}
Y.~Feng, J.~Lv, Y.~Cao, X.~Xie, and S.~K. Zhou, ``{Ada-KV}: Optimizing {KV}
  cache eviction by adaptive budget allocation for efficient {LLM} inference,''
  \emph{arXiv preprint arXiv:2407.11550}, 2024.

\bibitem{efficientcpt2025naacl}
A.~Nag, S.~Chakrabarti, A.~Mukherjee, and N.~Ganguly, ``Efficient continual
  pre-training of llms for low-resource languages,'' in \emph{Proceedings of
  the 2025 Conference of the Nations of the Americas Chapter of the Association
  for Computational Linguistics: Human Language Technologies (Volume 3:
  Industry Track)}, 2025.

\bibitem{agenticcontext2026}
Q.~Zhang, C.~Hu, S.~Upasani \emph{et~al.}, ``Agentic context engineering:
  Evolving contexts for self-improving language models,'' in \emph{The
  Fourteenth International Conference on Learning Representations}, 2026.

\bibitem{vaswani2017attention}
A.~Vaswani, N.~Shazeer, N.~Parmar \emph{et~al.}, ``Attention is all you need,''
  in \emph{Advances in Neural Information Processing Systems (NeurIPS)}, 2017.

\bibitem{brown2020language}
T.~B. Brown, B.~Mann, N.~Ryder \emph{et~al.}, ``Language models are few-shot
  learners,'' in \emph{Advances in Neural Information Processing Systems
  (NeurIPS)}, 2020.

\bibitem{yang2024qwen2}
A.~Yang, B.~Yang, B.~Hui \emph{et~al.}, ``Qwen2 technical report,'' \emph{arXiv
  preprint}, 2024.

\bibitem{qwen252024}
A.~Yang, B.~Yang, B.~Zhang \emph{et~al.}, ``Qwen2.5 technical report,''
  \emph{arXiv preprint}, 2025.

\bibitem{jiang2024mixtral}
A.~Q. Jiang, A.~Sablayrolles, A.~Roux \emph{et~al.}, ``Mixtral of experts,''
  \emph{arXiv preprint}, 2024.

\bibitem{deepseek2024v3}
D.~Guo, D.~Yang, H.~Zhang \emph{et~al.}, ``Deepseek-r1 incentivizes reasoning
  in llms through reinforcement learning,'' \emph{arXiv preprint}, 2025.

\bibitem{qwen32025}
A.~Yang, A.~Li, B.~Yang \emph{et~al.}, ``Qwen3 technical report,'' \emph{arXiv
  preprint}, 2025.

\bibitem{chen2021codex}
M.~Chen, J.~Tworek, H.~Jun \emph{et~al.}, ``Evaluating large language models
  trained on code,'' \emph{arXiv preprint}, 2021.

\bibitem{schick2023toolformer}
T.~Schick, J.~Dwivedi-Yu, R.~Dessi \emph{et~al.}, ``Toolformer: Language models
  can teach themselves to use tools,'' in \emph{Advances in Neural Information
  Processing Systems}, 2023.

\bibitem{wu2023autogen}
Q.~Wu, G.~Bansal, J.~Zhang \emph{et~al.}, ``Autogen: Enabling next-gen llm
  applications via multi-agent conversation,'' 2024.

\bibitem{yang2024sweagent}
J.~Yang, C.~Jimenez, A.~Wettig \emph{et~al.}, ``Swe-agent: Agent-computer
  interfaces enable automated software engineering,'' in \emph{Advances in
  Neural Information Processing Systems}, 2024.

\bibitem{anthropic2025claudecode}
Anthropic, ``Claude code: An agentic cli for software engineering,''
  \url{https://www.anthropic.com/news/claude-3-7-sonnet}, 2025, accessed:
  2026-05-18.

\bibitem{amini2019mathqa}
A.~Amini, S.~Gabriel, S.~Lin \emph{et~al.}, ``Mathqa: Towards interpretable
  math word problem solving with operation-based formalisms,'' in
  \emph{Proceedings of the 2019 Conference of the North {A}merican Chapter of
  the Association for Computational Linguistics: Human Language Technologies,
  Volume 1 (Long and Short Papers)}, 2019.

\bibitem{wang2024mathcoder}
K.~Wang, H.~Ren, A.~Zhou \emph{et~al.}, ``Mathcoder: Seamless code integration
  in llms for enhanced mathematical reasoning,'' in \emph{International
  Conference on Learning Representations}, 2024, pp. 5009--5042.

\bibitem{macina2023mathdial}
J.~Macina, N.~Daheim, S.~P. Chowdhury \emph{et~al.}, ``{M}ath{D}ial: A dialogue
  tutoring dataset with rich pedagogical properties grounded in math reasoning
  problems,'' in \emph{Findings of the Association for Computational
  Linguistics: EMNLP 2023}, 2023, pp. 5602--5621.

\bibitem{hinton2015distilling}
G.~Hinton, O.~Vinyals, and J.~Dean, ``Distilling the knowledge in a neural
  network,'' in \emph{Advances in Neural Information Processing Systems
  Workshop}, 2014.

\bibitem{gu2024minillm}
Y.~Gu, L.~Dong, F.~Wei, and M.~Huang, ``Minillm: Knowledge distillation of
  large language models,'' in \emph{International Conference on Learning
  Representations}, 2024, pp. 32\,694--32\,717.

\bibitem{hsieh2023distilling}
C.-Y. Hsieh, C.-L. Li, C.-K. Yeh \emph{et~al.}, ``Distilling step-by-step!
  outperforming larger language models with less training data and smaller
  model sizes,'' in \emph{Findings of the Association for Computational
  Linguistics: ACL 2023}, 2023, pp. 8003--8017.

\bibitem{chen2023extending}
S.~Chen, S.~Wong, L.~Chen, and Y.~Tian, ``Extending context window of large
  language models via positional interpolation,'' \emph{arXiv preprint}, 2023.

\bibitem{peng2024yarn}
B.~Peng, J.~Quesnelle, H.~Fan, and E.~Shippole, ``Yarn: Efficient context
  window extension of large language models,'' in \emph{International
  Conference on Learning Representations}, 2024, pp. 31\,932--31\,951.

\bibitem{dai2019transformerxl}
Z.~Dai, Z.~Yang, Y.~Yang \emph{et~al.}, ``Transformer-{XL}: Attentive language
  models beyond a fixed-length context,'' in \emph{Proceedings of the 57th
  Annual Meeting of the Association for Computational Linguistics}, 2019.

\bibitem{jiang2024minference}
H.~Jiang, Y.~Li, C.~Zhang \emph{et~al.}, ``Minference 1.0: Accelerating
  pre-filling for long-context llms via dynamic sparse attention,'' in
  \emph{Advances in Neural Information Processing Systems}, 2024, pp.
  52\,481--52\,515.

\bibitem{xiao2024efficient}
G.~Xiao, Y.~Tian, B.~Chen, S.~Han, and M.~Lewis, ``Efficient streaming language
  models with attention sinks,'' in \emph{International Conference on Learning
  Representations}, 2024, pp. 21\,875--21\,895.

\bibitem{han2024lminfinite}
C.~Han, Q.~Wang, H.~Peng \emph{et~al.}, ``Lm-infinite: Zero-shot extreme length
  generalization for large language models,'' in \emph{Proceedings of the 2024
  Conference of the North American Chapter of the Association for Computational
  Linguistics: Human Language Technologies (Volume 1: Long Papers)}, 2024, pp.
  3991--4008.

\bibitem{zhang2024h2o}
Z.~Zhang, Y.~Sheng, T.~Zhou \emph{et~al.}, ``H2o: Heavy-hitter oracle for
  efficient generative inference of large language models,'' in \emph{Advances
  in Neural Information Processing Systems}, 2023, pp. 34\,661--34\,710.

\bibitem{xiao2024infllm}
C.~Xiao, P.~Zhang, X.~Han \emph{et~al.}, ``Infllm: Training-free long-context
  extrapolation for llms with an efficient context memory,'' in \emph{Advances
  in Neural Information Processing Systems}, 2024, pp. 119\,638--119\,661.

\bibitem{sparseattention2024}
``Post-training sparse attention with double sparsity,'' \emph{arXiv preprint},
  2024.

\bibitem{sun2025shadowkv}
H.~Sun, L.-W. Chang, W.~Bao \emph{et~al.}, ``{ShadowKV}: {KV} cache in shadows
  for high-throughput long-context {LLM} inference,'' in \emph{Proceedings of
  the 42nd International Conference on Machine Learning}, ser. Proceedings of
  Machine Learning Research, vol. 267, 2025, pp. 57\,355--57\,373.

\bibitem{yang2024tidaldecode}
L.~Yang, Z.~Zhang, Z.~Chen, Z.~Li, and Z.~Jia, ``{TidalDecode}: Fast and
  accurate {LLM} decoding with position persistent sparse attention,'' in
  \emph{International Conference on Learning Representations (ICLR)}, 2025.

\bibitem{democratizing2024acl}
X.-P. Nguyen, M.~Aljunied, S.~Joty, and L.~Bing, ``Democratizing llms for
  low-resource languages by leveraging their english dominant abilities with
  linguistically-diverse prompts,'' in \emph{Proceedings of the 62nd Annual
  Meeting of the Association for Computational Linguistics (Volume 1: Long
  Papers)}, 2024.

\bibitem{ji2023survey}
Z.~Ji, N.~Lee, R.~Frieske \emph{et~al.}, ``Survey of hallucination in natural
  language generation,'' \emph{ACM Computing Surveys}, vol.~55, no.~12, pp.
  1--38, 2023.

\bibitem{huang2023survey}
L.~Huang, W.~Yu, W.~Ma \emph{et~al.}, ``A survey on hallucination in large
  language models: Principles, taxonomy, challenges, and open questions,''
  \emph{arXiv preprint arXiv:2311.05232}, 2023.

\bibitem{bang2023multitask}
Y.~Bang, S.~Cahyawijaya, N.~Lee \emph{et~al.}, ``A multitask, multilingual,
  multimodal evaluation of {ChatGPT} on reasoning, hallucination, and
  interactivity,'' in \emph{Proceedings of the 61st Annual Meeting of the
  Association for Computational Linguistics (Volume 1: Long Papers)}, 2023, pp.
  675--718.

\bibitem{promptalchemist2025}
S.~Gao, C.~Wang, C.~Gao \emph{et~al.}, ``The prompt alchemist: Automated
  llm-tailored prompt optimization for test case generation,'' \emph{arXiv
  preprint}, 2025.

\bibitem{krause2019dynamic}
B.~Krause, E.~Kahembwe, I.~Murray, and S.~Renals, ``Dynamic evaluation of
  transformer language models,'' \emph{arXiv preprint}, 2019.

\bibitem{shinn2023reflexion}
N.~Shinn, F.~Cassano, A.~Gopinath, K.~Narasimhan, and S.~Yao, ``Reflexion:
  language agents with verbal reinforcement learning,'' in \emph{Advances in
  Neural Information Processing Systems}, 2023, pp. 8634--8652.

\bibitem{nature2025optimizing}
M.~Yuksekgonul, F.~Bianchi, J.~Boen \emph{et~al.}, ``Optimizing generative ai
  by backpropagating language model feedback,'' \emph{Nature}, 2025.

\bibitem{gepa2025}
L.~A. Agrawal, S.~Tan, D.~Soylu \emph{et~al.}, ``Gepa: Reflective prompt
  evolution can outperform reinforcement learning,'' 2026.

\bibitem{memento_no_date}
H.~Zhou, Y.~Chen, S.~Guo \emph{et~al.}, ``Memento: Fine-tuning llm agents
  without fine-tuning llms,'' \emph{arXiv preprint}, 2025.

\bibitem{ainslie2023gqa}
J.~Ainslie, J.~Lee-Thorp, M.~de~Jong \emph{et~al.}, ``{GQA}: Training
  generalized multi-query transformer models from multi-head checkpoints,'' in
  \emph{Proceedings of the 2023 Conference on Empirical Methods in Natural
  Language Processing (EMNLP)}, 2023.

\bibitem{li2025scbench}
Y.~Li, H.~Jiang, Q.~Wu \emph{et~al.}, ``{SCBench}: A {KV} cache-centric
  analysis of long-context methods,'' in \emph{International Conference on
  Learning Representations (ICLR)}, 2025.

\bibitem{asl2026}
R.~Taniguchi, Y.~Dong, M.~Onizuka, and C.~Xiao, ``Adaptive layer selection for
  layer-wise token pruning in {LLM} inference,'' \emph{arXiv preprint
  arXiv:2601.07667}, 2026.

\bibitem{wu2025retrieval}
W.~Wu, Y.~Wang, G.~Xiao, H.~Peng, and Y.~Fu, ``Retrieval head mechanistically
  explains long-context factuality,'' in \emph{International Conference on
  Learning Representations (ICLR)}, 2025.

\bibitem{hao2026sparseengine}
J.~Hao, Q.~Gu, Q.~Huang, and J.~Yu, ``{SparseEngine}: Sparse-first inference
  engine,'' \emph{arXiv preprint arXiv:2609.39068}, 2026.

\bibitem{yang2025qwen251m}
A.~Yang, B.~Yu, C.~Li \emph{et~al.}, ``{Qwen2.5-1M} technical report,''
  \emph{arXiv preprint arXiv:2501.15383}, 2025.

\bibitem{zhang_etal_2024_bench}
X.~Zhang, Y.~Chen, S.~Hu \emph{et~al.}, ``$\infty$bench: Extending long context
  evaluation beyond 100k tokens,'' pp. 15\,262--15\,277, 2024.

\bibitem{ruler2024}
C.-P. Hsieh, S.~Sun, S.~Kriman \emph{et~al.}, ``Ruler: What's the real context
  size of your long-context language models?'' \emph{arXiv preprint}, 2024.

\bibitem{ester1996density}
M.~Ester, H.-P. Kriegel, J.~Sander, X.~Xu \emph{et~al.}, ``A density-based
  algorithm for discovering clusters in large spatial databases with noise,''
  in \emph{kdd}, vol.~96, no.~34, 1996, pp. 226--231.

\end{thebibliography}

\end{document}